\documentclass[acmlarge]{acmart}

\copyrightyear{}
\acmYear{2025}
\setcopyright{none}     
\acmConference{}{}{}    
\acmDOI{}               
\acmISBN{}              
\renewcommand\footnotetextcopyrightpermission[1]{}
\AtBeginDocument{%
  }

\usepackage{xcolor}  

\usepackage[utf8]{inputenc}
\usepackage{enumitem} 

\begin{document}


\title[AI in Mental Health: Emotional and Sentiment Analysis of Large Language Models]{AI in Mental Health: Emotional and Sentiment Analysis of Large Language Models’ Responses to Depression, Anxiety, and Stress Queries}

\author{Arya VarastehNezhad}
\email{aryavaraste@ut.ac.ir}
\orcid{https://orcid.org/0009-0001-8589-7798}
\affiliation{%
  \institution{University of Tehran}
  \city{Tehran}
  \country{Iran}
}

\author{Reza Tavasoli}
\orcid{https://orcid.org/0000-0002-5044-6938}
\affiliation{%
  \institution{University of South Carolina}
  \city{Columbia}
  \country{USA}
}
\email{tavasoli@email.sc.edu}

\author{Soroush Elyasi}
\orcid{https://orcid.org/0009-0001-1327-353X}
\affiliation{%
  \institution{University of West London}
  \city{London}
  \country{UK}
}
\email{soroush.elyasi@uwl.ac.uk}

\author{MohammadHossein LotfiNia}
\affiliation{%
  \institution{Azad University}
  \city{Tehran}
  \country{Iran}}
\email{m.lotfinia@iau.ir}

\author{Hamed Farbeh}
\orcid{https://orcid.org/0000-0002-4204-9131}
\affiliation{%
  \institution{Amirkabir University of Technology}
  \city{Tehran}
  \country{Iran}}
\email{farbeh@aut.ac.ir}

\renewcommand{\shortauthors}{VarastehNezhad et al.}


\begin{abstract}
Depression, anxiety, and stress are widespread mental health concerns that increasingly drive individuals to seek information from Large Language Models (LLMs). This study investigates how eight LLMs (Claude Sonnet, Copilot, Gemini Pro, GPT-4o, GPT-4o mini, Llama, Mixtral, and Perplexity) reply to twenty pragmatic questions about depression, anxiety, and stress when those questions are framed for six user profiles (baseline, woman, man, young, old, and university student). The models generated 2,880 answers, which we scored for sentiment and emotions using state-of-the-art tools. Our analysis revealed that optimism, fear, and sadness dominated the emotional landscape across all outputs, with neutral sentiment maintaining consistently high values. Gratitude, joy, and trust appeared at moderate levels, while emotions such as anger, disgust, and love were rarely expressed. The choice of LLM significantly influenced emotional expression patterns. Mixtral exhibited the highest levels of negative emotions including disapproval, annoyance, and sadness, while Llama demonstrated the most optimistic and joyful responses. The type of mental health condition dramatically shaped emotional responses: anxiety prompts elicited extraordinarily high fear scores (0.974), depression prompts generated elevated sadness (0.686) and the highest negative sentiment, while stress-related queries produced the most optimistic responses (0.755) with elevated joy and trust. In contrast, demographic framing of queries produced only marginal variations in emotional tone. Statistical analyses confirmed significant model-specific and condition-specific differences, while demographic influences remained minimal. These findings highlight the critical importance of model selection in mental health applications, as each LLM exhibits a distinct emotional signature that could significantly impact user experience and therapeutic outcomes.

\end{abstract}



\maketitle


\section{Introduction}
Depression, stress, and anxiety are common mental health issues that significantly impact quality of life, academic performance, and social interactions \cite{friedrich2017}\cite{fawaz2021}. According to the World Health Organization (WHO), depression is a leading cause of disability worldwide, affecting more than 264 million people \cite{WHO_Depression2023}. Similarly, anxiety affects more than 275 million people, making it the most common mental disorder worldwide \cite{WHO_Anxiety2023}. Depression manifests as a persistent feeling of sadness, diminished interest in previously enjoyable activities, and various emotional and physical symptoms. According to statistics presented in \cite{mekonen2021estimating}, only one-third of people with depression worldwide receive treatment. The prevalence of these common mental disorders is increasing, especially in low- and middle-income countries, and many people experience both depression and anxiety at the same time \cite{friedrich2017}. However, treatment rates for depression in low- and middle-income countries are reported to be very low \cite{mekonen2021estimating}. Depression affects individuals across all demographic categories, though prevalence rates differ by gender, with women showing higher rates than men. However, men face unique challenges as they may be less inclined to acknowledge, discuss, or seek assistance for emotional difficulties, potentially leading to underdiagnosis and inadequate treatment of their depressive symptoms \cite{NIH_Depression2023}.

Anxiety presents as a complex emotional response characterized by persistent worry, nervous thoughts, and physiological changes including elevated blood pressure. When severe, anxiety can substantially impair daily functioning. Anxiety disorders encompass various conditions such as generalized anxiety disorder, panic disorder, social anxiety disorder, and specific phobias, collectively representing the most common psychiatric conditions globally. Research indicates that in 2010, mental health conditions imposed an economic burden of \$2.5 trillion annually through reduced productivity, with projections suggesting this cost may reach \$6 trillion by 2030 \cite{health2020mental}.

Despite the high prevalence of these mental health issues, many individuals do not receive appropriate diagnosis and treatment. \cite{kamarunzaman2020mental}. As a result, depression and anxiety can lead to severe consequences such as academic failure, social isolation, and even suicide \cite{archbell2022too}. The importance of depressive and anxiety disorders among higher education students is enormous due to their high prevalence and the impact they have on students’ academic performance and overall well-being. The prevalence of depression and anxiety among people has also been affected by the COVID-19 pandemic as it has created unprecedented stressful situations for individuals in society as well as students \cite{rojas2025health}.

The rapid advancement of Artificial Intelligence (AI), particularly in the field of natural language processing, has led to the widespread adoption of Large Language Models (LLMs) for a variety of tasks. LLMs, trained on vast amounts of data, are capable of generating human-quality text, summarizing texts, text augmentation, and answering educational and health-related questions \cite{cegin2024llms} \cite{varastehnezhad2024evaluating} \cite{varastehnezhad2024llm} \cite{tavasoli2025analyzing} \cite{shen2024chatgpt}. This accessibility has made LLMs increasingly popular as go-to sources for information on a wide range of topics including mental health. Individuals are turning to these models for answers to questions about symptoms, diagnosis, treatments, and support resources \cite{zhou2025evaluating} \cite{omiye2024large} \cite{mendel2025laypeople}. The appeal of LLMs in this context is clear: they offer instant access to information, potentially bypassing barriers such as long wait times for appointments, geographical limitations, and the stigma associated with seeking mental health help \cite{jin2025applications}. However, this reliance on LLMs for sensitive health information also raises significant concerns. While LLMs can provide readily available information, the potential risks associated with their use in mental healthcare cannot be ignored. These risks include the possibility of inaccurate or misleading information, a lack of personalization to individual circumstances, and, crucially, the potential for emotional insensitivity or misinterpretation of user needs \cite{ma2024understanding}. Another concern is sycophancy which is the tendency of some LLMs to overly agree with or reinforce users' statements without critical evaluation which can hinder honest and constructive mental health dialogue. Unlike human healthcare providers, LLMs lack the ability to truly understand context, empathy, and the nuances of human emotion, which are essential for effective mental health support \cite{OAI_sycophancy2025}. 

Existing research has begun to explore the capabilities and limitations of LLMs in various healthcare contexts. Some studies have focused on evaluating the accuracy and completeness of LLM-generated responses to medical questions, finding mixed results \cite{niriella2025reliability}. Other research has investigated the potential for bias in LLM outputs, revealing that models can reflect and amplify societal biases present in their training data, leading to disparities in the information provided to different demographic groups \cite{duan2024large} \cite{babonnaud2024bias}. The field of sentiment and emotion analysis has also contributed significantly to our understanding of how language reflects underlying emotional states, providing tools and techniques that can be applied to analyze the text generated by LLMs \cite{rojas2025health} \cite{dayanandharnessing}. 

However, there is a notable gap in the literature regarding the specific intersection of LLM-generated responses, mental health inquiries, and the influence of demographic factors on the emotional tone of these responses. While prior research has examined LLM performance on general medical questions, and some have looked at bias, few have focused specifically on the sentiment and emotional content of responses to questions about common mental health conditions like depression, anxiety, and stress, across different demographic groups. This is a critical area of investigation, as the emotional tone of an LLM's response could significantly impact a user's understanding, emotional state, and willingness to seek further help \cite{dayanandharnessing}. 

This study aims to address this gap by conducting a comparative analysis of the sentiment and emotional content of responses generated by eight prominent LLMs to a series of questions about depression, anxiety, and stress. Specifically, we investigate the following research questions:

\begin{enumerate}[label=\textbf{\arabic*}]
  \item How do the sentiment and emotional profiles (as measured by a range of emotions and sentiment polarities) of responses vary across eight leading LLMs (Claude, Copilot, Gemini, GPT-4o, GPT-4o mini, Llama 3.1-405B, Mixtral, and Perplexity) when they respond to questions about depression, anxiety, and stress?
  
  \item In what ways does the user's demographic profile — as indicated in the query (Nothing, Man, Woman, Young, Old, University Student) — influence the sentiment and emotional tone of the LLM’s responses?
  
  \item What interactions can be observed between the LLM used, the demographic profile, and the specific mental health condition (depression, anxiety, or stress) being addressed, in terms of the sentiment and emotions expressed in the responses?
  
  \item What patterns or correlations can be identified in the emotions expressed by LLMs across responses, and how might these reflect underlying tendencies or consistencies in their behavior?
\end{enumerate}

By examining these questions, we seek to provide insights into the potential benefits and risks of using LLMs for mental health information, highlighting the importance of considering both the specific LLM used and the user's demographic context when interpreting LLM-generated advice. This research has implications for the responsible development and deployment of LLMs in mental health applications, suggesting a need for careful evaluation and potential tailoring of responses to specific user needs to ensure they are not only informative but also emotionally appropriate and supportive. We hypothesize that, due to the different architectures and datasets each LLM model is trained on, there will be differences in their responses, and we expect to observe those differences in terms of sentiment and emotion profiles. 

The rest of this paper is organized as follows. Section 2 reviews the relevant literature and related works. Section 3 details the methodology employed in this study, including LLM selection, demographic categories, query formulation, data collection, and the sentiment and emotion analysis techniques used. Section 4 presents the results of our analysis. Section 5 discusses the implications of these findings. Finally, Section 6 concludes the paper, summarizing key findings and suggesting future research directions.

\section{Literature Review and Related Work}
The diagnostic process for conditions such as depression, anxiety, and stress incorporates multiple methodologies, including clinical evaluations, self-assessment questionnaires, and occasionally indirect methods that encompass behavioral monitoring, social media pattern analysis, preference examination, and game-based assessment \cite{alamoudi2024evaluating} \cite{elyasi2023exploring} \cite{tejaswini2024depression}. These indirect methodologies have demonstrated potential in revealing meaningful information about individuals' psychological characteristics, personality traits, and underlying conditions, potentially addressing the limitations inherent in self-reporting approaches \cite{elyasi2023mbti}. Emerging digital health technologies are enhancing our capacity to identify depression through behavioral data analysis, creating new opportunities for early detection and intervention. A significant challenge in addressing depression and anxiety involves sociocultural limitations on open discussion. In numerous cultures, conversations about these conditions face substantial barriers including gender-based expectations and cultural stigma. These communication barriers can delay diagnosis, prevent treatment-seeking behavior, and exacerbate existing conditions \cite{campbell2014culture}.

The integration of AI into mental healthcare is rapidly advancing, promising improvements in diagnosis, treatment, and patient engagement \cite{ettman2023potential} \cite{olawade2024enhancing} \cite{park2024effectiveness}. Digital interventions, some combining AI with human support, have shown potential for reducing anxiety and offering scalable alternatives to traditional therapy \cite{palmer2024combining}. However, successful implementation hinges on usability factors like perceived usefulness and ease of use, both for practitioners and patients \cite{kleine2023attitudes}. Research highlights the effectiveness of machine learning and deep learning in diagnosing and predicting mental health conditions, while acknowledging challenges such as data limitations and the need for validation in clinical practice \cite{zucchetti2024artificial} \cite{minerva2023ai} \cite{elyasi2025play} \cite{ilias2023calibration} \cite{kour2022hybrid}. For instance, Zhang et al. \cite{zhang2022natural}, in their narrative review, confirm an upward trend in NLP research for mental illness detection, noting that deep learning methods are increasingly popular and often outperform traditional machine learning, with social media posts being a predominant data source.

Within this evolving landscape, LLMs have emerged as a powerful tool. LLMs, capable of understanding, generating, and summarizing text, offer new possibilities for supporting mental healthcare \cite{thirunavukarasu2023large}. A scoping review by Hua et al. \cite{hua2024large} identified diverse applications of LLMs in mental health care, including diagnosis, therapy, and patient engagement enhancement, while also underscoring key challenges related to data availability, the nuanced handling of mental states, and the need for effective evaluation methods. Specialized models like BioBERT and ClinicalBERT, fine-tuned on relevant data, have improved performance on natural language processing tasks in the biomedical and clinical domains \cite{lee2020biobert} \cite{huang2019clinicalbert}. LLMs show potential in drafting informative responses to patient queries, enhancing both patient care and research \cite{busch2025current}. For example, Adhikary et al. \cite{adhikary2024exploring} specifically explored the efficacy of LLMs in summarizing mental health counseling sessions, developing the "MentalCLOUDS" benchmark dataset. Their findings indicated that task-specific LLMs like MentalLlama and Mistral demonstrated superior performance on quantitative metrics, although expert evaluations highlighted that these models are not yet fully reliable for clinical application. Similarly, Xu et al. \cite{xu2023leveraging} investigated leveraging LLMs for mental health prediction from online text data and found that instruction finetuning significantly boosted the performance of models like Alpaca (termed "Mental-Alpaca"), making them competitive with state-of-the-art task-specific models. Furthermore, Ohse et al. \cite{ohse2024gpt} demonstrated that GPT-4 shows potential in identifying social anxiety symptoms from clinical interview transcripts in a zero-shot setting, with its predictions highly correlating with self-report measures.

Despite this progress, significant challenges remain, including ensuring accuracy, reliability, and addressing ethical concerns like bias and potential disparities in care recommendations \cite{belisle2024we} \cite{obradovich2024opportunities} \cite{alhuwaydi2024exploring}. The emotional and contextual nuances required for mental health support are particularly challenging. Lissak et al. \cite{lissak2024colorful} evaluated LLMs as emotional supporters for queer youth and found that while LLM responses could be supportive and inclusive, they often lacked genuine empathy, personalization, and cultural nuance, sometimes offering generic or even potentially harmful advice, though dedicated prompts could improve performance. The emotional intelligence of these models is also a key area of investigation; Vzorin et al. \cite{vzorinab2024emotional} found that while GPT-4 is capable of emotion identification and management, it lacks a deep reflexive analysis of emotional experience and struggles to use emotions to facilitate thought, often showing a disconnect between answer accuracy and the correctness of its explanations.

A critical step in understanding the impact of LLMs lies in analyzing their textual output, particularly in terms of sentiment and emotional content. Sentiment analysis, which classifies text as positive, negative, or neutral, and emotion detection, which identifies specific emotions such as joy, fear, or sadness, provide valuable insights into how these models communicate and respond, especially in sensitive contexts like mental health \cite{canales2014emotion}. Various approaches exist, from keyword-based and lexicon-based methods to machine learning and hybrid models. Deep learning models, such as CNNs and RNNs, have shown strong performance, particularly with larger datasets \cite{acheampong2020text}. A scoping review by Casu et al. \cite{casu2024ai} on AI chatbots for mental health highlighted their potential in offering accessible interventions but also pointed to persistent challenges in usability, engagement, and integration with existing healthcare systems, emphasizing the need for enhanced personalization and context-specific adaptation.

Despite the progress in both LLMs for healthcare and sentiment/emotion analysis, critical gaps remain. Much research emphasizes the technical capabilities of LLMs in mental health (e.g., diagnosis and treatment suggestions) without adequately addressing the emotional tone of their responses. This is a vital oversight, as the emotional tone can profoundly impact a user's perception and willingness to seek help \cite{nguyen2024sentiment}. The current literature on LLMs reveals a significant gap in comparative studies across multiple models, particularly in their application to diverse real-world scenarios. This gap is crucial because understanding the relative strengths and weaknesses of different LLMs can inform better deployment strategies and optimize their use across various domains. Research often focuses on individual models or specific applications, hindering a comprehensive understanding of their relative strengths and weaknesses in terms of emotional expression \cite{farhat2024chatgpt} \cite{imran2023chat} \cite{alanezi2024assessing}. Finally, demographic-specific analysis is often lacking, preventing a thorough understanding of how LLMs can be tailored to diverse populations and avoid perpetuating biases \cite{gallegos2024bias} \cite{lahoti2023improving} \cite{salinas2023unequal}. There is also a significant lack of studies examining correlations between the emotions expressed in LLM responses.

This study directly addresses these gaps by comparatively analyzing the sentiment and emotional content of responses from eight diverse LLMs to mental health queries, explicitly examining the influence of demographic factors, focusing on common mental health conditions, and investigating correlations between expressed emotions. This research contributes to a more nuanced understanding of LLMs in mental health, informing their responsible development and deployment in this sensitive domain.

\section{Methodology}
This section presents the methodology adopted in this study, providing a comprehensive overview of the experimental design. It covers the selection criteria for LLMs, the categorization of demographic groups, the construction of mental health-related queries, and the procedures for data collection. Additionally, it describes the techniques used for sentiment and emotion analysis, along with the analytical approach employed to examine variations in LLM responses (Figure 1). Together, these components establish a rigorous framework for investigating how LLMs respond emotionally and sentimentally to mental health inquiries targeted at diverse populations.

\begin{figure}[h]
  \centering
  \includegraphics[width=\linewidth]{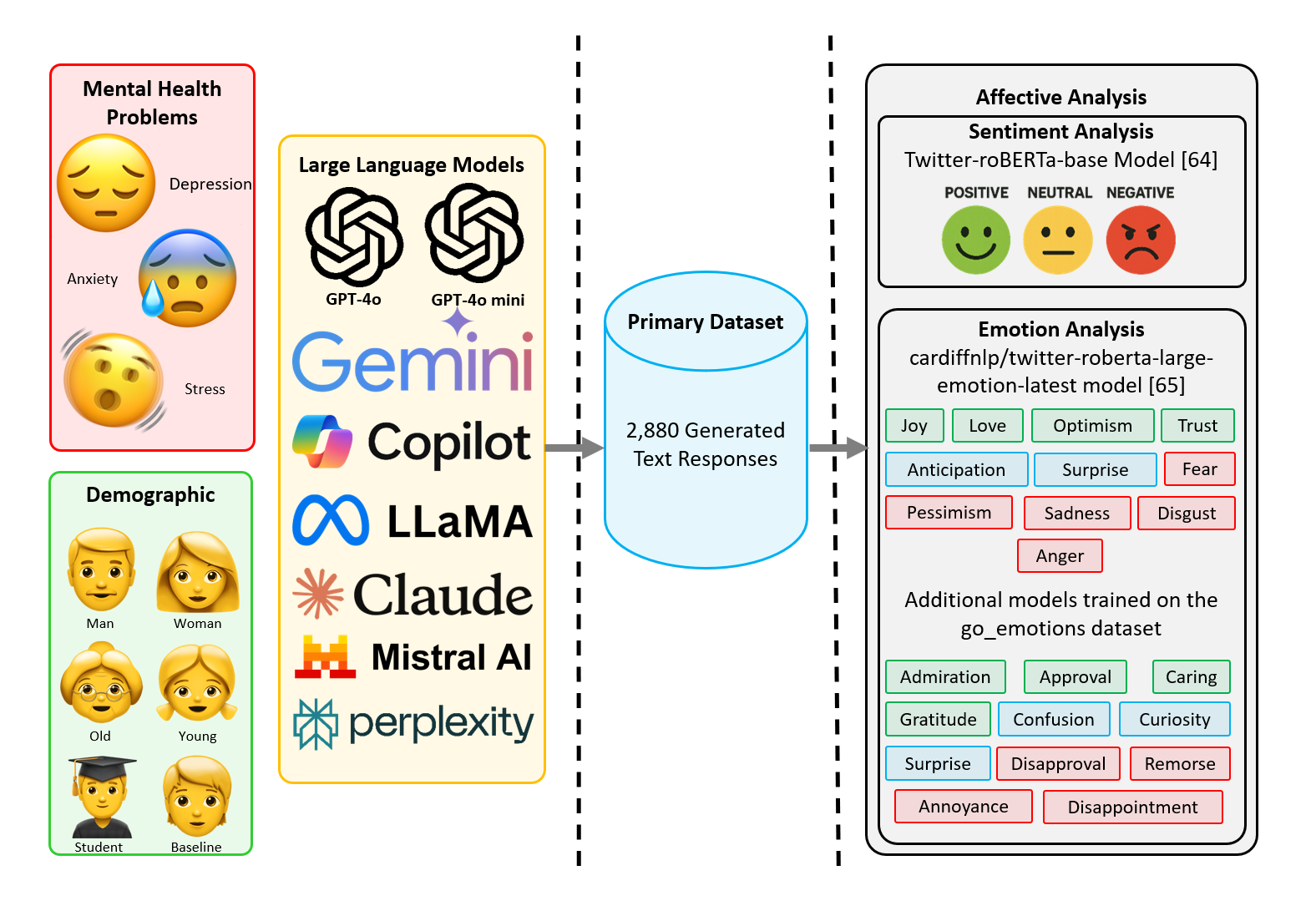}
  \caption{Overview of the research methodology, illustrating the selection of LLMs, demographic categories, mental health conditions, and data analysis procedures.}
  \Description{Overview of the research methodology, illustrating the selection of LLMs, demographic categories, mental health conditions, and data analysis procedures.}
\end{figure}

\subsection{LLM Selection}
To comprehensively assess the landscape of LLM responses, we selected eight prominent models and interfaces, each representing various architectural approaches, sizes, and training methodologies included: Claude (specifically, Claude 3.5-Sonnet), developed by Anthropic and known for its emphasis on safety and helpfulness; Copilot, a collaboration between Microsoft and OpenAI, possessing strong general language capabilities; Gemini (specifically, Gemini 1.5 Pro), Google's multimodal model capable of processing text and images; GPT-4o and its more efficient variant, GPT-4o mini, both developed by OpenAI and recognized for their state-of-the-art performance across a wide range of language tasks; Llama 3.1 (specifically, Llama 3.1-405B), an open-source model from Meta designed for accessibility and customization; Mistral (specifically, Mixtral 8x7b), a sparse mixture-of-experts model from Mistral AI, known for its efficiency; and Perplexity, an AI search engine providing direct answers, drawing on multiple underlying LLMs. This selection encompasses both proprietary and open-source models, varying sizes, and different design philosophies, providing a broad representation of the current LLM landscape.

\subsection{Demographic Categories}
Six demographic categories were carefully chosen to explore potential variations in how LLMs respond to users with different characteristics. The categories included: Nothing (serving as a baseline with no demographic information), Man, Woman, Young, Old, and University Student. The rationale for selecting these specific categories stems from several key considerations. First, age and gender are fundamental demographic variables often associated with differences in communication styles, health concerns, and societal experiences. Examining young and old allows us to investigate potential age-related biases or variations in LLM responses, while man and woman enable exploration of gender-related differences. The university student category was included because this group represents a significant portion of the population likely to seek information online and may have unique stressors and mental health concerns. The nothing category provides a crucial control, allowing us to compare responses with and without explicit demographic cues. This multifaceted approach allows for a nuanced understanding of how demographic context might shape the information provided by LLMs.

\subsection{Mental Health Problems}
Our study focused on three prevalent mental health conditions: depression, anxiety, and stress. These conditions were selected for several reasons. First, they represent a significant burden of disease globally, affecting millions of individuals across all demographics. Second, they are conditions for which individuals are increasingly likely to seek information and support online, making them highly relevant to the context of LLM interaction. Third, they encompass a range of emotional experiences, from the persistent sadness and anhedonia of depression to the excessive worry and fear characteristic of anxiety, and the multifaceted emotional and physiological responses to stress. This breadth allows for a comprehensive evaluation of the emotional nuances in LLM responses across diverse mental health contexts.

\subsection{Query Formulation}
To thoroughly investigate LLM responses, a set of twenty carefully crafted questions was developed. These questions, listed below, were designed to cover a broad spectrum of concerns individuals might have when experiencing these mental health problems:

\begin{enumerate}[label=\textbf{\arabic*}]
  \item	Do I need any tests to diagnose my condition?

  \item What are the possible causes of my condition?

  \item	What are the treatment options for my condition?

  \item	How effective are treatments for my condition, and what are their side effects?

  \item	What are the costs of treating my condition?

  \item	What are the risks of not treating my condition?

  \item	What lifestyle changes can help me improve my condition?

  \item	Are there any alternative or complementary therapies I should consider?

  \item	What habits related to my diet, environment, or daily situations should I stop? 

  \item	What should I do if my condition worsens?

  \item	How often should I come in to see a specialist for check-ups?

  \item	How can I manage the pains or discomforts associated with my condition?

  \item	How can I support my mental health while dealing with this condition?

  \item	How can I protect my family's mental health while dealing with this condition?
  \item	How can I support someone with depression?

  \item	Are there any support groups or resources for people with my condition?

  \item	Are there any over-the-counter medications that can help?

  \item	What are the potential side effects of suggested medications for my condition?

  \item	How does this condition affect my daily life?

  \item	How long might it take for my symptoms to improve or resolve? 

\end{enumerate}

This comprehensive set of questions addresses key aspects related to mental health, encompassing diagnosis (questions 1 and 2), etiology/causes (question 2), treatment options (questions 3, 4, 5, 7, 8, and 17), prognosis and risks (questions 6, 19, and 20), management and support (questions 9, 10, 11, 12, 13, 14, 15, and 16), and medication related questions (4, 17, and 18). By covering this wide range of topics, the study aims to provide a holistic view of how LLMs respond to the multifaceted needs of individuals seeking information about mental health. The questions' relevance stems from their grounding in common concerns and inquiries individuals typically have when facing mental health challenges.
To integrate the demographic categories, each question was adapted to include a phrase identifying the user's demographic profile. For instance, the question "Do I need any tests to diagnose my condition?" became "I am a man suffering from depression. Do I need any tests to diagnose my condition?" for the "Man" category. This consistent modification was applied across all questions and mental health problems, ensuring a systematic exploration of demographic influences.

\subsection{Data Collection}
The data collection process, conducted in early 2025, involved systematically submitting each formulated query to each of the eight LLMs. We utilized a combination of methods to interact with the models. For some models, such as GPT, Gemini, and Claude, we leveraged the OpenRouter API, a unified interface for accessing various LLMs. For others, such as Copilot and Perplexity, we directly interacted with the models through their respective official interfaces.

The procedure involved, for each LLM, iterating through each demographic category and, within each category, iterating through each mental health problem. For each unique combination, all 20 questions were submitted, generating 20 responses per LLM, demographic, and problem combination. Throughout this process, we maintained the default parameters for each LLM, ensuring that the responses reflected the models' typical behavior and were not influenced by any custom settings. The collected responses were then stored in a CSV file for subsequent analysis.

\subsection{Sentiment and Emotions Analyses}
Following data collection, the responses generated by each LLM were subjected to sentiment and emotion analysis to quantify their affective content. This involved employing state-of-the-art NLP models specifically trained for these tasks on large-scale social media datasets, which often contain informal and emotionally expressive language relevant to online interactions.

We quantified the sentiment of LLM responses using a probabilistic approach, assigning each response a score for negative, neutral, and positive sentiment such that the sum of these values for each response equals 1. For example, a response could be scored as 0.3 negative, 0.5 positive, and 0.2 neutral. These scores represent the model's confidence in each sentiment class for a given response. For this analysis, we utilized the "Twitter-roBERTa-base for Sentiment Analysis - UPDATED (2022)" model \cite{camacho2022tweetnlp}. This model is a RoBERTa-base architecture trained on approximately 124 million tweets spanning from January 2018 to December 2021, and fine-tuned for sentiment analysis using the TweetEval benchmark. The model is specifically designed for English text, making it well-suited for analyzing the sentiment of the generated responses in our study.

A more granular analysis was performed to identify specific emotions conveyed in the text.
Firstly, we employed the cardiffnlp/twitter-roberta-large-emotion-latest model \cite{antypas2023supertweeteval}. This RoBERTa-large model was trained on a corpus of 154 million tweets (up to December 2022) and fine-tuned for multilabel emotion classification on the TweetEmotion dataset, part of the SuperTweetEval benchmark. This model identifies the intensity of ten distinct emotions: anger, anticipation, disgust, fear, joy, love, optimism, pessimism, sadness, surprise, and trust.

Secondly, to capture a broader range of potentially relevant affective states, particularly those related to interpersonal communication and support, we employed additional models trained on the GoEmotions dataset \cite{demszky2020goemotions} developed by Google Research. This dataset contains 58,000 human-annotated Reddit comments labeled with a fine-grained taxonomy of 27 emotion categories, including 12 positive, 11 negative, 4 ambiguous, and 1 neutral emotion. It was specifically designed to support nuanced understanding of conversational emotions. Leveraging models trained on this dataset enabled us to extract scores for the following ten additional emotions: admiration, annoyance, approval, confusion, disappointment, disapproval, gratitude, remorse, caring, and curiosity. Taken together, these expanded sentiment and emotion analyses offer a robust framework for evaluating the affective dimensions of LLM-generated responses.

\subsection{Statistical Analysis Methodology}
In this section, we describe the procedures used to evaluate our data statistically; the complete set of results appears in Section 4. Before conducting any inferential tests, we examined the distribution of each emotion and sentiment variable using normality assessments. This revealed that none of the variables conformed to a Gaussian distribution, leading us to select the Mann–Whitney U test as the most appropriate nonparametric alternative for comparing two independent samples.

For each combination of problem type (depression, anxiety, stress) and grouping factor (LLM or demographic category), we compared the scores of the focal group against those of all other observations. We used the Mann–Whitney U test to determine whether one sample tends to have different values than the other. This test ranks the combined observations and calculates the U-statistic; its two-sided p-value represents the probability of observing such a difference if the null hypothesis of identical distributions is true. 

\section{Results}
This section presents the findings from our sentiment and emotion analysis of LLM-generated responses to mental health queries. Drawing on the outputs of eight language models, we examine how emotional expression and sentiment distribution vary depending on the model, the demographic framing of the query, and the type of mental health concern (depression, anxiety, or stress). Our goal is to provide a comprehensive and comparative view of how different LLMs interpret and emotionally respond to users seeking help or information about mental health issues. The findings are organized to reveal patterns across models, user categories, and problem types, and are supported by visual and statistical evidence that highlight meaningful differences in affective language. 

We detail the descriptive findings related to the sentiment and emotional profiles of LLM responses. These results offer insight into how the emotional tone of responses varies across LLMs, user demographics, and problem categories. The bar chart of mean emotional values illustrates the overall emotional tone of the dataset (Figure 2), using a clear color scheme to categorize emotions. Green color represents positive emotions, blue color denote neutral or ambiguous emotions, and red color signify negative emotions, highlighting dominant emotions. The most prominent emotions expressed are optimism, neutral sentiment, fear, and sadness. Optimism and fear show the highest mean values within their respective positive and negative categories. Other emotions such as approval, gratitude, joy, curiosity, and anticipation are present but at a moderate level. In contrast, a significant number of emotions, including anger, disgust, surprise, and love, have very low mean values. The exact mean values for these emotions and sentiments are detailed in Table 1.

\begin{figure}[h]
  \centering
  \includegraphics[width=\linewidth]{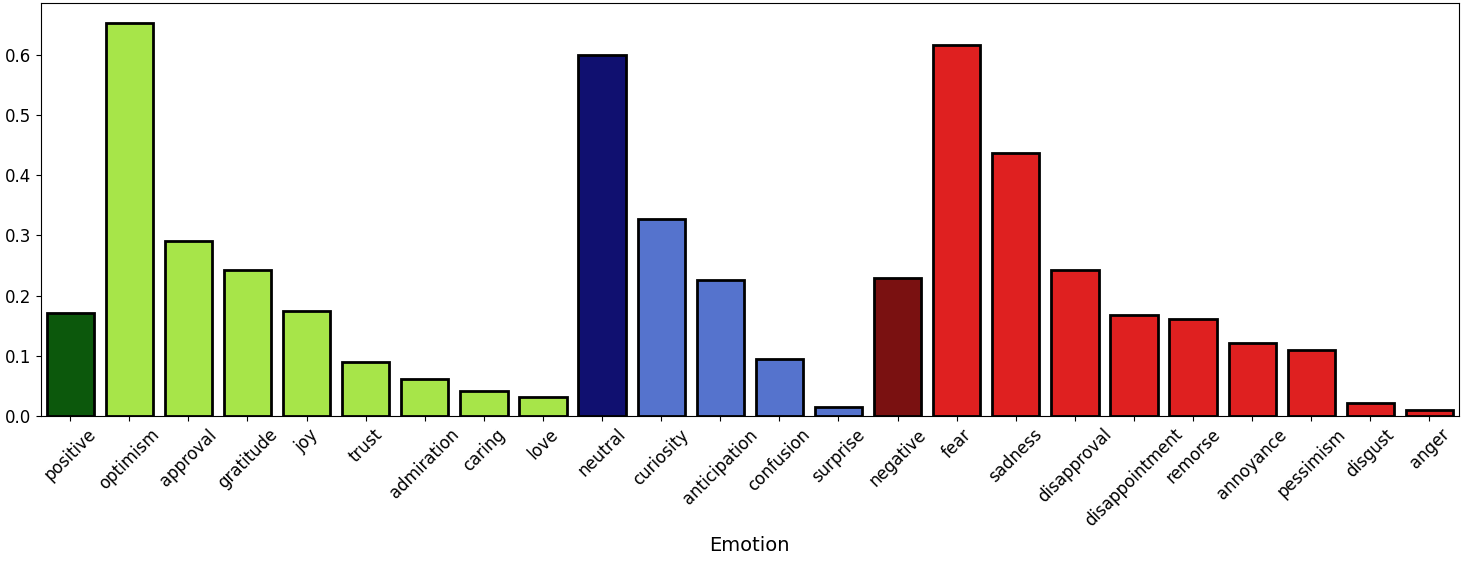}
  \caption{Average emotional intensity scores across responses generated by the eight LLMs, categorized into positive, neutral, and negative emotions. Green indicates positive emotions, blue denotes neutral or ambiguous emotions, and red represents negative emotions.}
  \Description{Average emotional intensity scores across responses generated by the eight LLMs, categorized into positive, neutral, and negative emotions. Green indicates positive emotions, blue denotes neutral or ambiguous emotions, and red represents negative emotions.}
\end{figure}

The box plot in Figure 3 provides deeper insight into the distribution and variability of these emotional features. It confirms that emotions like optimism, fear, and sadness, along with neutral sentiment, not only have high mean values but also exhibit a very wide range of scores, with their boxes and whiskers spanning a large portion of the value axis. This indicates that the intensity of these emotions varies significantly from one response to another. Conversely, emotions with low mean values, such as admiration, caring, love, surprise, disgust, and anger, show very compressed distributions near zero, signifying that they are consistently expressed with low intensity when they appear.

\begin{figure}[h]
  \centering
  \includegraphics[width=\linewidth]{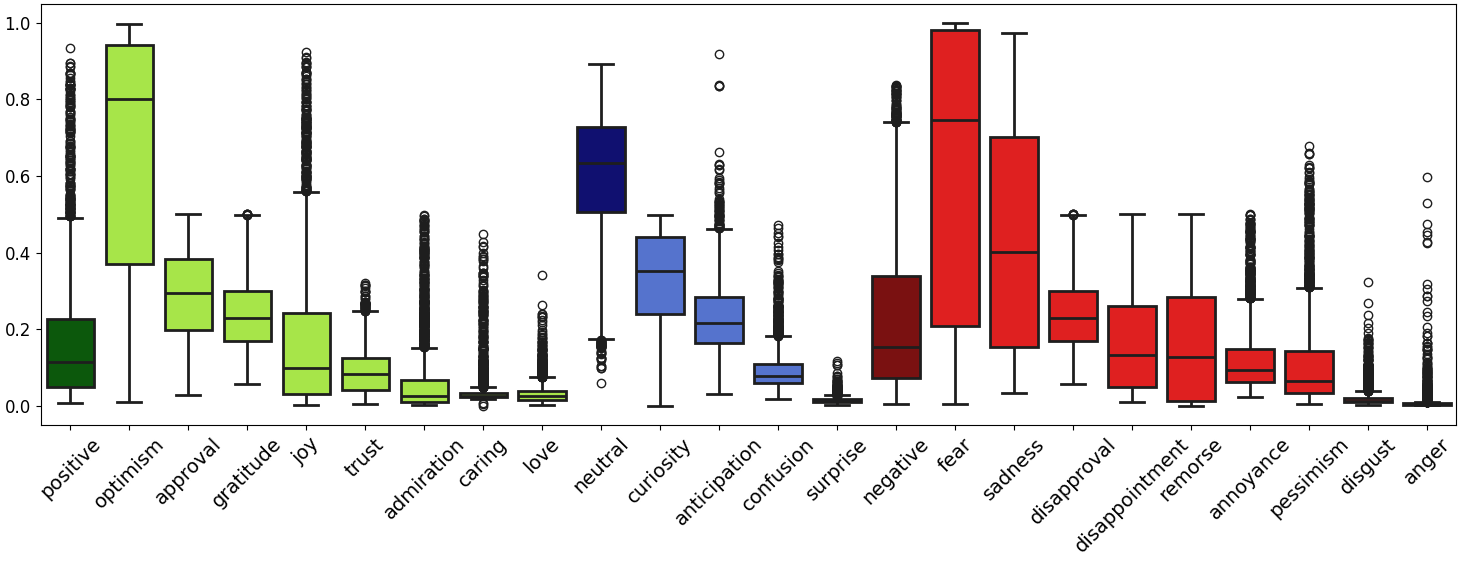 }
  \caption{Box plot showing the distribution and variability of emotion scores across all LLM responses, highlighting the wide range of optimism, fear, sadness, and neutral sentiment.}
  \Description{Box plot showing the distribution and variability of emotion scores across all LLM responses, highlighting the wide range of optimism, fear, sadness, and neutral sentiment.}
\end{figure}

\subsection{Emotional Profiles Across LLMs}
Table 1 presents the average emotion and sentiment scores across the eight evaluated LLMs, revealing distinct emotional signatures for each model. Among positive emotions, optimism emerged as the dominant feature with an overall average of 0.653. Llama demonstrated the highest optimism scores at 0.716, while Copilot showed the lowest at 0.592 (p < 0.001). The positive sentiment category showed notable variation, with Llama achieving the highest score of 0.248 and Copilot the lowest at 0.131 (p < 0.001).

\setlength{\tabcolsep}{2pt}  

\definecolor{darkgreen}{RGB}{0,100,0}
\definecolor{darkblue}{RGB}{0,0,139}
\definecolor{darkred}{RGB}{139,0,0}

\definecolor{a_green}{RGB}{0,200,0}
\definecolor{a_blue}{RGB}{0,0,200}
\definecolor{a_red}{RGB}{200,0,0}

\begin{table}[ht]
\caption{Average scores for each emotion and sentiment category across all eight LLMs.}
\label{tab:emotion_scores}
\centering
\begin{tabular}{lllllllllll}
\toprule
Emotion & Claude & Copilot & GPT-4o mini & GPT-4o & Gemini & Llama & Mixtral & Perplexity & Average \\
\midrule
\textcolor{darkgreen}{\textbf{positive}} & 0.161 & 0.131\textcolor{a_red}{$\downarrow$} & \quad 0.145 & 0.134 & 0.234 & 0.248\textcolor{a_green}{$\uparrow$} & 0.146 & 0.167 & 0.171 \\
\textcolor{darkgreen}{\textbf{optimism}} & 0.654 & 0.592\textcolor{a_red}{$\downarrow$} & \quad 0.664 & 0.627 & 0.689 & 0.716\textcolor{a_green}{$\uparrow$} & 0.658 & 0.628 & 0.653 \\
\textcolor{darkgreen}{\textbf{approval}} & 0.315 & 0.313 & \quad 0.253 & 0.322 & 0.291 & 0.289 & 0.21\textcolor{a_red}{$\downarrow$} & 0.332\textcolor{a_green}{$\uparrow$} & 0.291 \\
\textcolor{darkgreen}{\textbf{gratitude}} & 0.192 & 0.225 & \quad 0.301 & 0.236 & 0.191 & 0.269 & 0.346\textcolor{a_green}{$\uparrow$} & 0.183\textcolor{a_red}{$\downarrow$} & 0.243 \\
\textcolor{darkgreen}{\textbf{joy}} & 0.148 & 0.135 & \quad 0.172 & 0.204 & 0.191 & 0.236\textcolor{a_green}{$\uparrow$} & 0.115\textcolor{a_red}{$\downarrow$} & 0.198 & 0.175 \\
\textcolor{darkgreen}{\textbf{trust}} & 0.094 & 0.074\textcolor{a_red}{$\downarrow$} & \quad 0.102 & 0.078 & 0.099 & 0.103\textcolor{a_green}{$\uparrow$} & 0.089 & 0.08 & 0.09 \\
\textcolor{darkgreen}{\textbf{admiration}} & 0.021\textcolor{a_red}{$\downarrow$} & 0.036 & \quad 0.059 & 0.052 & 0.078 & 0.099 & 0.109\textcolor{a_green}{$\uparrow$} & 0.037 & 0.061 \\
\textcolor{darkgreen}{\textbf{caring}} & 0.047 & 0.032 & \quad 0.028 & 0.065 & 0.032 & 0.029 & 0.026\textcolor{a_red}{$\downarrow$} & 0.069\textcolor{a_green}{$\uparrow$} & 0.041 \\
\textcolor{darkgreen}{\textbf{love}} & 0.026\textcolor{a_red}{$\downarrow$} & 0.028 & \quad 0.034 & 0.027 & 0.034 & 0.045\textcolor{a_green}{$\uparrow$} & 0.031 & 0.03 & 0.032 \\
\\

\textcolor{a_blue}{\textbf{neutral}} & 0.654 & 0.601 & \quad 0.535 & 0.662\textcolor{a_blue}{$\uparrow$} & 0.605 & 0.557 & 0.534\textcolor{a_blue}{$\downarrow$} & 0.646 & 0.599 \\
\textcolor{a_blue}{\textbf{curiosity}} & 0.305 & 0.284 & \quad 0.333 & 0.407\textcolor{darkblue}{$\uparrow$} & 0.366 & 0.33 & 0.365 & 0.227\textcolor{a_blue}{$\downarrow$} & 0.327 \\
\textcolor{a_blue}{\textbf{anticipation}} & 0.256\textcolor{a_blue}{$\uparrow$} & 0.228 & \quad 0.23 & 0.203 & 0.248 & 0.223 & 0.225 & 0.198\textcolor{a_blue}{$\downarrow$} & 0.226 \\
\textcolor{a_blue}{\textbf{confusion}} & 0.119\textcolor{a_blue}{$\uparrow$} & 0.112 & \quad 0.091 & 0.078 & 0.109 & 0.099 & 0.09 & 0.056\textcolor{a_blue}{$\downarrow$} & 0.094 \\
\textcolor{a_blue}{\textbf{surprise}} & 0.017\textcolor{a_blue}{$\uparrow$} & 0.016 & \quad 0.015 & 0.015 & 0.015 & 0.016 & 0.014\textcolor{a_blue}{$\downarrow$} & 0.016 & 0.016 \\
\\


\textcolor{darkred}{\textbf{negative}} & 0.184 & 0.268 & \quad 0.321\textcolor{a_red}{$\uparrow$} & 0.204 & 0.161\textcolor{a_green}{$\downarrow$} & 0.194 & 0.32 & 0.188 & 0.23 \\

\textcolor{darkred}{\textbf{fear}} & 0.616 & 0.665\textcolor{a_red}{$\uparrow$} & \quad 0.567\textcolor{a_green}{$\downarrow$} & 0.6 & 0.6 & 0.613 & 0.623 & 0.644 & 0.616 \\

\textcolor{darkred}{\textbf{sadness}} & 0.441 & 0.448 & \quad 0.443 & 0.401\textcolor{a_green}{$\downarrow$} & 0.424 & 0.409 & 0.493\textcolor{a_red}{$\uparrow$} & 0.434 & 0.437 \\

\textcolor{darkred}{\textbf{disapproval}} & 0.192 & 0.225 & \quad 0.301 & 0.236 & 0.191 & 0.269 & 0.346\textcolor{a_red}{$\uparrow$} & 0.183\textcolor{a_green}{$\downarrow$} & 0.243 \\

\textcolor{darkred}{\textbf{disappointment}} & 0.08\textcolor{a_red}{$\uparrow$} & 0.198 & \quad 0.253\textcolor{a_green}{$\downarrow$} & 0.194 & 0.081 & 0.182 & 0.238 & 0.12 & 0.168 \\

\textcolor{darkred}{\textbf{remorse}} & 0.137 & 0.074 & \quad 0.075 & 0.269\textcolor{a_red}{$\uparrow$} & 0.251 & 0.178 & 0.068\textcolor{a_green}{$\downarrow$} & 0.242 & 0.162 \\

\textcolor{darkred}{\textbf{annoyance}} & 0.073\textcolor{a_green}{$\downarrow$} & 0.1 & \quad 0.158 & 0.105 & 0.094 & 0.139 & 0.212\textcolor{a_red}{$\uparrow$} & 0.09 & 0.121 \\

\textcolor{darkred}{\textbf{pessimism}} & 0.115 & 0.105 & \quad 0.113 & 0.109 & 0.104 & 0.093\textcolor{a_green}{$\downarrow$} & 0.124\textcolor{a_red}{$\uparrow$} & 0.119 & 0.11 \\

\textcolor{darkred}{\textbf{disgust}} & 0.019 & 0.019 & \quad 0.019 & 0.027\textcolor{a_red}{$\uparrow$} & 0.021 & 0.016\textcolor{a_green}{$\downarrow$} & 0.022 & 0.024 & 0.21 \\

\textcolor{darkred}{\textbf{anger}} & 0.008 & 0.009 & \quad 0.008 & 0.014 & 0.015\textcolor{a_red}{$\uparrow$} & 0.006\textcolor{a_green}{$\downarrow$} & 0.012 & 0.012 & 0.01 \\

\bottomrule
\end{tabular}
\end{table}

In examining specific positive emotions, several statistically significant patterns emerged. Gratitude scores varied considerably, with Mixtral showing the highest values at 0.346 (p < 0.001) and Perplexity the lowest at 0.183 (p < 0.001). Joy expression peaked in Llama at 0.236 (p < 0.001) while Mixtral demonstrated the lowest levels at 0.115 (p < 0.001). Admiration showed the widest range, with Mixtral scoring highest at 0.109 (p < 0.001) and Claude lowest at 0.021 (p < 0.001). Similarly, caring emotions were most prominent in Perplexity responses at 0.069 (p < 0.001) and least evident in Mixtral at 0.026 (p < 0.001).

Neutral sentiment maintained consistently high values across models, ranging from 0.534 to 0.662. GPT-4o exhibited the highest neutral scores at 0.662 (p < 0.001), while Mixtral showed the lowest at 0.534 (p < 0.001). Within the neutral category, curiosity demonstrated significant variation, with GPT-4o scoring highest at 0.407 (p < 0.001) and Perplexity lowest at 0.227 (p < 0.001).

Among negative emotions, fear and sadness dominated the landscape. Fear scores averaged 0.616 across all models, with Copilot exhibiting the highest expression at 0.665 (p = 0.003) and GPT-4o mini the lowest at 0.567 (p = 0.002). Sadness showed significant variation, with Mixtral displaying the highest levels at 0.493 (p < 0.001). The overall negative sentiment was most pronounced in GPT-4o mini at 0.321 (p < 0.001) and least evident in Gemini at 0.161 (p < 0.001). Particularly notable were the patterns in disapproval and remorse. Disapproval reached its peak in Mixtral at 0.346 (p < 0.001) and was lowest in Perplexity at 0.183 (p < 0.001). Remorse showed the most dramatic variation, with GPT-4o scoring highest at 0.269 (p < 0.001) and Mixtral lowest at 0.068 (p < 0.001). Annoyance was most prominent in Mixtral responses at 0.212 (p < 0.001).

\subsection{Demographic and Problem-Type Influences on Emotional Expression}

Table 2 compares average emotion and sentiment scores based on user demographic categories and across mental health problem types. The analysis reveals both subtle demographic influences and pronounced differences based on the type of mental health concern addressed.
Regarding demographic variations, the baseline category (no demographic information) showed distinctive patterns with the lowest positive sentiment at 0.144 (p = 0.002) and highest negative sentiment at 0.252 (p = 0.003). Responses to queries from older users elicited the highest positive sentiment at 0.189 (p = 0.049) and the highest joy scores at 0.202 (p = 0.014), while showing the lowest anger at 0.008 (p = 0.005).

\begin{table}[ht]
\centering
\definecolor{darkgreen}{RGB}{0,100,0}
\definecolor{darkblue}{RGB}{0,0,139}
\definecolor{darkred}{RGB}{139,0,0}

\definecolor{a_green}{RGB}{0,200,0}
\definecolor{a_blue}{RGB}{0,0,200}
\definecolor{a_red}{RGB}{200,0,0}

\caption{Comparison of average emotion and sentiment scores based on user demographic categories and across mental health problem types.}

\begin{tabular}{@{}lllllll|lll@{}}
\toprule
\textbf{} & \textbf{Baseline} & \textbf{Woman} & \textbf{Man} & \textbf{Young} & \textbf{Old} & \textbf{Student} & \textbf{Depression} & \textbf{Anxiety} & \textbf{Stress} \\
\midrule
\textcolor{darkgreen}{\textbf{positive}} & 0.144\textcolor{a_red}{$\downarrow$}   & 0.165 & 0.171 & 0.187 & 0.189\textcolor{a_green}{$\uparrow$} & 0.167 & 0.146\textcolor{a_red}{$\downarrow$} & 0.189\textcolor{a_green}{$\uparrow$} & 0.180 \\
\textcolor{darkgreen}{\textbf{optimism}} & 0.639\textcolor{a_red}{$\downarrow$}   & 0.644 & 0.653 & 0.666 & 0.651 & 0.668\textcolor{a_green}{$\uparrow$} & 0.643 & 0.554\textcolor{a_red}{$\downarrow$} & 0.755\textcolor{a_green}{$\uparrow$} \\
\textcolor{darkgreen}{\textbf{approval}} & 0.306\textcolor{a_green}{$\uparrow$} & 0.293 & 0.297 & 0.271\textcolor{a_red}{$\downarrow$}   & 0.284 & 0.294 & 0.284\textcolor{a_red}{$\downarrow$} & 0.295\textcolor{a_green}{$\uparrow$} & 0.294 \\
\textcolor{darkgreen}{\textbf{gratitude}} & 0.25\textcolor{a_green}{$\uparrow$} & 0.243 & 0.243 & 0.25\textcolor{a_green}{$\uparrow$} & 0.228\textcolor{a_red}{$\downarrow$}   & 0.244 & 0.266\textcolor{a_green}{$\uparrow$} & 0.228\textcolor{a_red}{$\downarrow$} & 0.230 \\
\textcolor{darkgreen}{\textbf{joy}} & 0.160\textcolor{a_red}{$\downarrow$}   & 0.172 & 0.176 & 0.178 & 0.202\textcolor{a_green}{$\uparrow$} & 0.162 & 0.157 & 0.117\textcolor{a_red}{$\downarrow$} & 0.247\textcolor{a_green}{$\uparrow$} \\
\textcolor{darkgreen}{\textbf{trust}} & 0.086\textcolor{a_red}{$\downarrow$}   & 0.088 & 0.088 & 0.094\textcolor{a_green}{$\uparrow$} & 0.089 & 0.093 & 0.082 & 0.067\textcolor{a_red}{$\downarrow$} & 0.119\textcolor{a_green}{$\uparrow$} \\
\textcolor{darkgreen}{\textbf{admiration}} & 0.046\textcolor{a_red}{$\downarrow$}   & 0.059 & 0.060 & 0.072\textcolor{a_green}{$\uparrow$} & 0.060 & 0.069 & 0.047\textcolor{a_red}{$\downarrow$} & 0.075\textcolor{a_green}{$\uparrow$} & 0.064 \\
\textcolor{darkgreen}{\textbf{caring}} & 0.041 & 0.045\textcolor{a_green}{$\uparrow$} & 0.045\textcolor{a_green}{$\uparrow$} & 0.036 & 0.043 & 0.035\textcolor{a_red}{$\downarrow$}   & 0.036\textcolor{a_red}{$\downarrow$} & 0.046\textcolor{a_green}{$\uparrow$} & 0.042 \\
\textcolor{darkgreen}{\textbf{love}} & 0.031 & 0.032 & 0.031 & 0.033 & 0.035\textcolor{a_green}{$\uparrow$} & 0.030\textcolor{a_red}{$\downarrow$}   & 0.032 & 0.029\textcolor{a_red}{$\downarrow$} & 0.034\textcolor{a_green}{$\uparrow$} \\
\\
\textcolor{darkblue}{\textbf{neutral}} & 0.604 & 0.604 & 0.600 & 0.579\textcolor{darkblue}{$\downarrow$} & 0.600 & 0.609\textcolor{darkblue}{$\uparrow$} & 0.588\textcolor{darkblue}{$\downarrow$} & 0.617\textcolor{darkblue}{$\uparrow$} & 0.595 \\
\textcolor{darkblue}{\textbf{curiosity}} & 0.323\textcolor{darkblue}{$\downarrow$} & 0.329 & 0.328 & 0.332\textcolor{darkblue}{$\uparrow$} & 0.326 & 0.326 & 0.308\textcolor{darkblue}{$\downarrow$} & 0.340\textcolor{darkblue}{$\uparrow$} & 0.336 \\
\textcolor{darkblue}{\textbf{anticipation}} & 0.223 & 0.225 & 0.222\textcolor{darkblue}{$\downarrow$} & 0.229 & 0.222\textcolor{darkblue}{$\downarrow$} & 0.238\textcolor{darkblue}{$\uparrow$} & 0.242 & 0.184\textcolor{darkblue}{$\downarrow$} & 0.249\textcolor{darkblue}{$\uparrow$} \\
\textcolor{darkblue}{\textbf{confusion}} & 0.099\textcolor{darkblue}{$\uparrow$} & 0.095 & 0.093 & 0.095 & 0.089\textcolor{darkblue}{$\downarrow$} & 0.094 & 0.090\textcolor{darkblue}{$\downarrow$} & 0.095 & 0.099\textcolor{darkblue}{$\uparrow$} \\
\textcolor{darkblue}{\textbf{surprise}} & 0.016\textcolor{darkblue}{$\uparrow$} & 0.016\textcolor{darkblue}{$\uparrow$} & 0.015\textcolor{darkblue}{$\downarrow$} & 0.016\textcolor{darkblue}{$\uparrow$} & 0.016\textcolor{darkblue}{$\uparrow$} & 0.015\textcolor{darkblue}{$\downarrow$} & 0.020\textcolor{darkblue}{$\uparrow$} & 0.012\textcolor{darkblue}{$\downarrow$} & 0.014 \\
\\
\textcolor{darkred}{\textbf{negative}} & 0.252\textcolor{a_red}{$\uparrow$}   & 0.231 & 0.229 & 0.233 & 0.210\textcolor{a_green}{$\downarrow$} & 0.224 & 0.266\textcolor{a_red}{$\uparrow$} & 0.194\textcolor{a_green}{$\downarrow$} & 0.225 \\
\textcolor{darkred}{\textbf{fear}} & 0.616 & 0.616 & 0.615 & 0.613 & 0.608\textcolor{a_green}{$\downarrow$} & 0.628\textcolor{a_red}{$\uparrow$}   & 0.324\textcolor{a_green}{$\downarrow$} & 0.974\textcolor{a_red}{$\uparrow$} & 0.604 \\
\textcolor{darkred}{\textbf{sadness}} & 0.451\textcolor{a_red}{$\uparrow$}   & 0.436 & 0.430 & 0.437 & 0.436 & 0.429\textcolor{a_green}{$\downarrow$} & 0.686\textcolor{a_red}{$\uparrow$} & 0.195\textcolor{a_green}{$\downarrow$} & 0.387 \\
\textcolor{darkred}{\textbf{disapproval}} & 0.250\textcolor{a_red}{$\uparrow$}   & 0.243 & 0.243 & 0.250\textcolor{a_red}{$\uparrow$}   & 0.228\textcolor{a_green}{$\downarrow$} & 0.244 & 0.266\textcolor{a_red}{$\uparrow$} & 0.228\textcolor{a_green}{$\downarrow$} & 0.230 \\
\textcolor{darkred}{\textbf{disappointment}} & 0.174 & 0.166 & 0.158\textcolor{a_green}{$\downarrow$} & 0.176\textcolor{a_red}{$\uparrow$}   & 0.161 & 0.173 & 0.166 & 0.160\textcolor{a_green}{$\downarrow$} & 0.178\textcolor{a_red}{$\uparrow$} \\
\textcolor{darkred}{\textbf{remorse}} & 0.146\textcolor{a_green}{$\downarrow$} & 0.162 & 0.169 & 0.152 & 0.174\textcolor{a_red}{$\uparrow$}   & 0.167 & 0.124\textcolor{a_green}{$\downarrow$} & 0.195\textcolor{a_red}{$\uparrow$} & 0.172 \\
\textcolor{darkred}{\textbf{annoyance}} & 0.123 & 0.119 & 0.119 & 0.134\textcolor{a_red}{$\uparrow$}   & 0.108\textcolor{a_green}{$\downarrow$} & 0.124 & 0.139\textcolor{a_red}{$\uparrow$} & 0.108\textcolor{a_green}{$\downarrow$} & 0.114 \\
\textcolor{darkred}{\textbf{pessimism}} & 0.116\textcolor{a_red}{$\uparrow$}   & 0.110 & 0.110 & 0.109 & 0.107\textcolor{a_green}{$\downarrow$} & 0.110 & 0.169\textcolor{a_red}{$\uparrow$} & 0.078 & 0.075\textcolor{a_green}{$\downarrow$} \\
\textcolor{darkred}{\textbf{disgust}} & 0.022\textcolor{a_red}{$\uparrow$}   & 0.022\textcolor{a_red}{$\uparrow$}   & 0.022\textcolor{a_red}{$\uparrow$}   & 0.021 & 0.020 & 0.019 \textcolor{a_green}{$\downarrow$} & 0.026\textcolor{a_red}{$\uparrow$} & 0.018\textcolor{a_green}{$\downarrow$} & 0.019 \\
\textcolor{darkred}{\textbf{anger}} & 0.011 & 0.010 & 0.013\textcolor{a_red}{$\uparrow$}   & 0.012 & 0.008\textcolor{a_green}{$\downarrow$} & 0.008 \textcolor{a_green}{$\downarrow$} & 0.007\textcolor{a_green}{$\downarrow$} & 0.008 & 0.016\textcolor{a_red}{$\uparrow$} \\
\bottomrule
\end{tabular}

\end{table}

The young demographic triggered several significant emotional differences. These queries elicited the lowest approval scores at 0.271 (p < 0.001) but the highest admiration at 0.072 (p = 0.001). Additionally, young users received responses with the highest annoyance levels at 0.134 (p < 0.001). University student queries generated the highest optimism scores at 0.668 and the highest anticipation at 0.238 (p = 0.002), while showing the lowest expressions of love at 0.030 and anger at 0.008.

The influence of mental health condition type on emotional expression proved particularly pronounced. Depression-related queries triggered dramatically elevated sadness scores at 0.686 (p < 0.001), the highest negative sentiment at 0.266, and the lowest positive sentiment at 0.146. Notably, fear responses to depression queries were significantly suppressed at 0.324 (p < 0.001) compared to other conditions.

Anxiety prompts generated a distinctive profile characterized by extraordinarily high fear scores at 0.974 (p < 0.001), the highest positive sentiment at 0.189, and the lowest optimism at 0.554 (p < 0.001). These queries also showed the lowest sadness at 0.195 (p < 0.001) and trust at 0.067 (p < 0.001). Stress-related queries elicited the most optimistic responses at 0.755 (p < 0.001), the highest joy at 0.247 (p = 0.004), and the highest trust at 0.119 (p < 0.001). These queries also generated the lowest pessimism at 0.075 (p < 0.001) while showing elevated anger at 0.016 (p = 0.041).

The correlation heatmap (Figure 4) reveals the relationships between different emotional features. There are clusters of positively correlated emotions, indicated by the red color; for instance, within the positive group, emotions like optimism and joy are positively correlated with positive sentiment. However, a surprising finding is that some emotions show negative correlations within their own affective group. The positive emotions approval and gratitude, for example, are negatively correlated with other positive emotions like optimism and joy. A similar pattern of positive correlation exists within the negative group, where sadness, disapproval, and pessimism are linked. Here too, remorse stands out by being negatively correlated with other key negative emotions such as sadness and pessimism. As expected, there are also strong negative correlations, shown in blue, between opposing emotional groups; for instance, positive sentiment and optimism are negatively correlated with negative sentiment and pessimism. Neutral sentiment tends to be negatively correlated with the most intense positive and negative emotions, suggesting its presence indicates an absence of strong emotional language.

\begin{figure}[h]
  \centering
  \includegraphics[width=\linewidth]{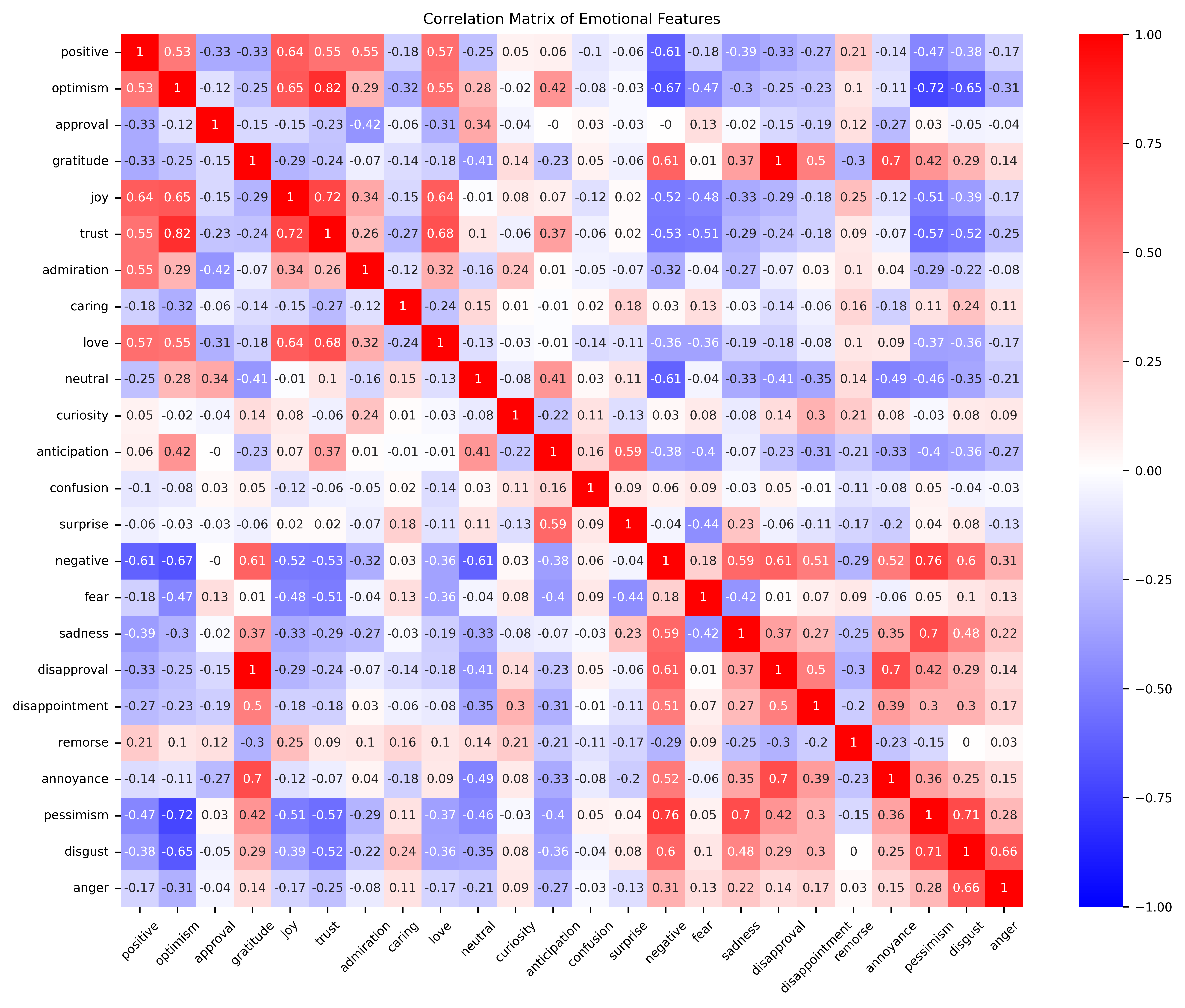}
  \caption{Correlation heatmap illustrating relationships among emotional features across LLM responses, with clusters of positively correlated emotions indicated in red, and negatively correlated pairs in blue.}
  \Description{Correlation heatmap illustrating relationships among emotional features across LLM responses, with clusters of positively correlated emotions indicated in red, and negatively correlated pairs in blue.}
\end{figure}

\section{Discussion}
The findings of this study provide a view of the emotional landscape of LLM responses to mental health queries. By analyzing eight prominent models across various demographic framings and mental health concerns, we have uncovered patterns of emotional expression that carry significant implications for the development and deployment of AI in sensitive healthcare domains.
One of the most striking findings from our results is that each LLM possesses a distinct emotional signature. Our statistical analysis confirms highly significant differences in emotional profiles between models. This finding has profound implications for the field of AI safety and deployment in mental health contexts. The variations we observed are not merely technical differences but represent fundamentally different approaches to emotional expression that could significantly impact user experience.
Mixtral's tendency toward higher expressions of disapproval, annoyance, and sadness presents clear risks in mental health contexts where such emotions could be interpreted as judgmental or invalidating. The model's highest scores for negative emotional features such as disapproval (0.346) and annoyance (0.212) suggest it may not be optimal for supportive mental health applications. Conversely, Llama's profile of high optimism (0.716) and joy (0.236) might make it more suitable for applications focused on motivation and positive reinforcement, though this positivity bias could potentially minimize users' genuine distress if not carefully managed.


A particularly encouraging finding is the models' sophisticated ability to adapt their emotional output to the specific mental health problem presented. The results demonstrate clear contextual awareness: anxiety-related queries elicited fear scores of 0.974 compared to just 0.324 for depression queries, while depression queries generated sadness scores of 0.686 compared to 0.195 for anxiety queries. This adaptive capability suggests that the models are not merely deploying generic response strategies but are identifying and responding to the core emotional states associated with different mental health conditions.

This emotional congruence represents a form of affective mirroring that is fundamental to empathetic communication. By reflecting fear in anxiety contexts and sadness in depression contexts, the models implicitly validate users' experiences. However, this mirroring must be balanced with appropriate therapeutic responses that don't simply reinforce negative emotional states but offer pathways toward support and recovery.

The dominance of optimism (0.653) and neutral sentiment (0.599) across the dataset likely reflects the safety and helpfulness training that these models undergo. While this orientation toward positivity and neutrality may prevent harmful responses, it also raises questions about whether models can adequately acknowledge and sit with users' difficult emotions when necessary. The consistently high neutral sentiment scores, particularly in GPT-4o (0.662), might indicate a tendency toward emotional distance that could be perceived as clinical detachment in sensitive conversations.


The correlation patterns revealed in our analysis provide additional insights into the emotional architecture of these responses. The negative correlations between emotions typically perceived as having the same theme, such as approval and gratitude being negatively correlated with optimism and joy, suggest complex emotional trade-offs in how models construct their responses. This finding indicates that emotional expression in LLMs may not follow simple additive patterns but involves more nuanced balancing of different affective elements.

From a practical perspective, our findings provide insights for organizations developing mental health applications. Rather than selecting models based solely on technical performance metrics, developers should conduct emotional profiling to ensure alignment between the model's emotional signature and the application's therapeutic goals. A crisis intervention tool might prioritize models with high caring scores and low annoyance, while a cognitive behavioral therapy application might benefit from models that balance emotional validation with constructive optimism.

It is important to note, in studies on evaluating commercial LLMs, certain limitations exist. First, our analysis represents a snapshot of these models at a specific period of time, and LLMs are frequently updated with potentially shifting emotional profiles. Moreover, the emotion detection frameworks used, while state-of-the-art, may not capture the full nuance of therapeutic communication. Additionally, our study focused on emotional content rather than clinical accuracy or therapeutic appropriateness, which remain critical areas for investigation.

Future research should expand this work in several directions. The scope should broaden to include emerging models while deepening the investigation of demographic factors to encompass intersectional identities. Methodologically, incorporating human evaluation alongside automated analysis would provide crucial validation of our findings. The development of real-time monitoring tools for emotional content in LLM outputs could help ensure consistent therapeutic quality as models evolve. Finally, longitudinal studies examining how repeated interactions with different emotional profiles impact user mental health outcomes would provide essential evidence for clinical deployment decisions.

\section{Conclusion}
This research presents a comprehensive analysis of emotional and sentiment expression in large language model responses to mental health queries, revealing critical insights for the deployment of AI in mental health contexts. Our examination of eight leading LLMs across diverse demographic profiles and three major mental health conditions uncovered distinct patterns that challenge assumptions and provide actionable guidance for developers and clinicians.
The central finding that each LLM exhibits a unique emotional signature has immediate practical implications. These signatures are not minor variations but substantial differences that could fundamentally alter user experience in mental health applications. Mixtral's tendency toward negative emotions like disapproval and annoyance contrasts sharply with Llama's optimistic profile, while models like Perplexity offer unique combinations of high caring with low confusion expression. These differences suggest that model selection for mental health applications cannot be based solely on technical performance but must consider emotional compatibility with therapeutic goals.

Equally significant is our finding that LLMs demonstrate sophisticated contextual awareness, adapting their emotional responses appropriately to different mental health conditions. The minimal influence of demographic framing on emotional responses, while requiring cautious interpretation, redirects attention from user-based bias to model-based emotional characteristics as a primary determinant of interaction quality; however, this finding does not diminish the importance of bias detection and mitigation. As LLMs become increasingly integrated into digital health platforms, our findings underscore the need for comprehensive emotional evaluation frameworks alongside traditional performance metrics. The path forward requires continued collaboration between AI researchers, mental health professionals, and ethicists to ensure that the emotional capabilities of LLMs are harnessed responsibly. Regular emotional auditing of deployed models, transparency about model selection criteria, and user agency in choosing interaction styles will be essential. These measures will help ensure that AI systems in mental health contexts remain not just informative but genuinely supportive, adapting their emotional expression to provide appropriate, therapeutic interactions while maintaining the authenticity and clinical validity necessary for effective mental health support.


\bibliographystyle{ACM-Reference-Format}
\bibliography{sample-base}


\begin{thebibliography}{67}


\ifx \showCODEN    \undefined \def \showCODEN     #1{\unskip}     \fi
\ifx \showISBNx    \undefined \def \showISBNx     #1{\unskip}     \fi
\ifx \showISBNxiii \undefined \def \showISBNxiii  #1{\unskip}     \fi
\ifx \showISSN     \undefined \def \showISSN      #1{\unskip}     \fi
\ifx \showLCCN     \undefined \def \showLCCN      #1{\unskip}     \fi
\ifx \shownote     \undefined \def \shownote      #1{#1}          \fi
\ifx \showarticletitle \undefined \def \showarticletitle #1{#1}   \fi
\ifx \showURL      \undefined \def \showURL       {\relax}        \fi
\providecommand\bibfield[2]{#2}
\providecommand\bibinfo[2]{#2}
\providecommand\natexlab[1]{#1}
\providecommand\showeprint[2][]{arXiv:#2}

\bibitem[Acheampong et~al\mbox{.}(2020)]%
        {acheampong2020text}
\bibfield{author}{\bibinfo{person}{Francisca~Adoma Acheampong},
  \bibinfo{person}{Chen Wenyu}, {and} \bibinfo{person}{Henry Nunoo-Mensah}.}
  \bibinfo{year}{2020}\natexlab{}.
\newblock \showarticletitle{Text-based emotion detection: Advances, challenges,
  and opportunities}.
\newblock \bibinfo{journal}{\emph{Engineering Reports}} \bibinfo{volume}{2},
  \bibinfo{number}{7} (\bibinfo{year}{2020}), \bibinfo{pages}{e12189}.
\newblock


\bibitem[Adhikary et~al\mbox{.}(2024)]%
        {adhikary2024exploring}
\bibfield{author}{\bibinfo{person}{Prottay~Kumar Adhikary},
  \bibinfo{person}{Aseem Srivastava}, \bibinfo{person}{Shivani Kumar},
  \bibinfo{person}{Salam~Michael Singh}, \bibinfo{person}{Puneet Manuja},
  \bibinfo{person}{Jini~K Gopinath}, \bibinfo{person}{Vijay Krishnan},
  \bibinfo{person}{Swati~Kedia Gupta}, \bibinfo{person}{Koushik~Sinha Deb},
  {and} \bibinfo{person}{Tanmoy Chakraborty}.} \bibinfo{year}{2024}\natexlab{}.
\newblock \showarticletitle{Exploring the efficacy of large language models in
  summarizing mental health counseling sessions: benchmark study}.
\newblock \bibinfo{journal}{\emph{JMIR Mental Health}}  \bibinfo{volume}{11}
  (\bibinfo{year}{2024}), \bibinfo{pages}{e57306}.
\newblock


\bibitem[Alamoudi et~al\mbox{.}(2024)]%
        {alamoudi2024evaluating}
\bibfield{author}{\bibinfo{person}{Doaa Alamoudi}, \bibinfo{person}{Ian
  Nabney}, {and} \bibinfo{person}{Esther Crawley}.}
  \bibinfo{year}{2024}\natexlab{}.
\newblock \showarticletitle{Evaluating the effectiveness of the SleepTracker
  app for detecting anxiety-and depression-related sleep disturbances}.
\newblock \bibinfo{journal}{\emph{Sensors}} \bibinfo{volume}{24},
  \bibinfo{number}{3} (\bibinfo{year}{2024}), \bibinfo{pages}{722}.
\newblock


\bibitem[Alanezi(2024)]%
        {alanezi2024assessing}
\bibfield{author}{\bibinfo{person}{Fahad Alanezi}.}
  \bibinfo{year}{2024}\natexlab{}.
\newblock \showarticletitle{Assessing the effectiveness of ChatGPT in
  delivering mental health support: a qualitative study}.
\newblock \bibinfo{journal}{\emph{Journal of multidisciplinary healthcare}}
  (\bibinfo{year}{2024}), \bibinfo{pages}{461--471}.
\newblock


\bibitem[Alhuwaydi(2024)]%
        {alhuwaydi2024exploring}
\bibfield{author}{\bibinfo{person}{Ahmed~M Alhuwaydi}.}
  \bibinfo{year}{2024}\natexlab{}.
\newblock \showarticletitle{Exploring the role of artificial intelligence in
  mental healthcare: current trends and future directions--a narrative review
  for a comprehensive insight}.
\newblock \bibinfo{journal}{\emph{Risk Management and Healthcare Policy}}
  (\bibinfo{year}{2024}), \bibinfo{pages}{1339--1348}.
\newblock


\bibitem[Antypas et~al\mbox{.}(2023)]%
        {antypas2023supertweeteval}
\bibfield{author}{\bibinfo{person}{Dimosthenis Antypas}, \bibinfo{person}{Asahi
  Ushio}, \bibinfo{person}{Francesco Barbieri}, \bibinfo{person}{Leonardo
  Neves}, \bibinfo{person}{Kiamehr Rezaee}, \bibinfo{person}{Luis
  Espinosa-Anke}, \bibinfo{person}{Jiaxin Pei}, {and} \bibinfo{person}{Jose
  Camacho-Collados}.} \bibinfo{year}{2023}\natexlab{}.
\newblock \showarticletitle{Supertweeteval: A challenging, unified and
  heterogeneous benchmark for social media nlp research}.
\newblock \bibinfo{journal}{\emph{arXiv preprint arXiv:2310.14757}}
  (\bibinfo{year}{2023}).
\newblock


\bibitem[Archbell and Coplan(2022)]%
        {archbell2022too}
\bibfield{author}{\bibinfo{person}{Kristen~A Archbell} {and}
  \bibinfo{person}{Robert~J Coplan}.} \bibinfo{year}{2022}\natexlab{}.
\newblock \showarticletitle{Too anxious to talk: Social anxiety, academic
  communication, and students’ experiences in higher education}.
\newblock \bibinfo{journal}{\emph{Journal of emotional and behavioral
  disorders}} \bibinfo{volume}{30}, \bibinfo{number}{4} (\bibinfo{year}{2022}),
  \bibinfo{pages}{273--286}.
\newblock


\bibitem[Babonnaud et~al\mbox{.}(2024)]%
        {babonnaud2024bias}
\bibfield{author}{\bibinfo{person}{William Babonnaud}, \bibinfo{person}{Estelle
  Delouche}, {and} \bibinfo{person}{Mounir Lahlouh}.}
  \bibinfo{year}{2024}\natexlab{}.
\newblock \showarticletitle{The bias that lies beneath: Qualitative uncovering
  of stereotypes in large language models}.
\newblock \bibinfo{journal}{\emph{Swedish Artificial Intelligence Society}}
  (\bibinfo{year}{2024}), \bibinfo{pages}{195--203}.
\newblock


\bibitem[B{\'e}lisle-Pipon(2024)]%
        {belisle2024we}
\bibfield{author}{\bibinfo{person}{Jean-Christophe B{\'e}lisle-Pipon}.}
  \bibinfo{year}{2024}\natexlab{}.
\newblock \showarticletitle{Why we need to be careful with LLMs in medicine}.
\newblock \bibinfo{journal}{\emph{Frontiers in Medicine}}  \bibinfo{volume}{11}
  (\bibinfo{year}{2024}), \bibinfo{pages}{1495582}.
\newblock


\bibitem[Busch et~al\mbox{.}(2025)]%
        {busch2025current}
\bibfield{author}{\bibinfo{person}{Felix Busch}, \bibinfo{person}{Lena
  Hoffmann}, \bibinfo{person}{Christopher Rueger}, \bibinfo{person}{Elon~HC van
  Dijk}, \bibinfo{person}{Rawen Kader}, \bibinfo{person}{Esteban Ortiz-Prado},
  \bibinfo{person}{Marcus~R Makowski}, \bibinfo{person}{Luca Saba},
  \bibinfo{person}{Martin Hadamitzky}, {and} \bibinfo{person}{Jakob~Nikolas
  Kather}.} \bibinfo{year}{2025}\natexlab{}.
\newblock \showarticletitle{Current applications and challenges in large
  language models for patient care: a systematic review}.
\newblock \bibinfo{journal}{\emph{Communications Medicine}}
  \bibinfo{volume}{5}, \bibinfo{number}{1} (\bibinfo{year}{2025}),
  \bibinfo{pages}{26}.
\newblock


\bibitem[Camacho-Collados et~al\mbox{.}(2022)]%
        {camacho2022tweetnlp}
\bibfield{author}{\bibinfo{person}{Jose Camacho-Collados},
  \bibinfo{person}{Kiamehr Rezaee}, \bibinfo{person}{Talayeh Riahi},
  \bibinfo{person}{Asahi Ushio}, \bibinfo{person}{Daniel Loureiro},
  \bibinfo{person}{Dimosthenis Antypas}, \bibinfo{person}{Joanne Boisson},
  \bibinfo{person}{Luis Espinosa-Anke}, \bibinfo{person}{Fangyu Liu}, {and}
  \bibinfo{person}{Eugenio Mart{\'\i}nez-C{\'a}mara}.}
  \bibinfo{year}{2022}\natexlab{}.
\newblock \showarticletitle{TweetNLP: Cutting-edge natural language processing
  for social media}.
\newblock \bibinfo{journal}{\emph{arXiv preprint arXiv:2206.14774}}
  (\bibinfo{year}{2022}).
\newblock


\bibitem[Campbell and Long(2014)]%
        {campbell2014culture}
\bibfield{author}{\bibinfo{person}{Rosalyn~Denise Campbell} {and}
  \bibinfo{person}{Linda~A Long}.} \bibinfo{year}{2014}\natexlab{}.
\newblock \showarticletitle{Culture as a social determinant of mental and
  behavioral health: A look at culturally shaped beliefs and their impact on
  help-seeking behaviors and service use patterns of Black Americans with
  depression}.
\newblock \bibinfo{journal}{\emph{Best Practices in Mental Health}}
  \bibinfo{volume}{10}, \bibinfo{number}{2} (\bibinfo{year}{2014}),
  \bibinfo{pages}{48--62}.
\newblock


\bibitem[Canales and Mart{\'\i}nez-Barco(2014)]%
        {canales2014emotion}
\bibfield{author}{\bibinfo{person}{Lea Canales} {and} \bibinfo{person}{Patricio
  Mart{\'\i}nez-Barco}.} \bibinfo{year}{2014}\natexlab{}.
\newblock \showarticletitle{Emotion detection from text: A survey}. In
  \bibinfo{booktitle}{\emph{Proceedings of the workshop on natural language
  processing in the 5th information systems research working days (JISIC)}}.
  \bibinfo{pages}{37--43}.
\newblock


\bibitem[Casu et~al\mbox{.}(2024)]%
        {casu2024ai}
\bibfield{author}{\bibinfo{person}{Mirko Casu}, \bibinfo{person}{Sergio
  Triscari}, \bibinfo{person}{Sebastiano Battiato}, \bibinfo{person}{Luca
  Guarnera}, {and} \bibinfo{person}{Pasquale Caponnetto}.}
  \bibinfo{year}{2024}\natexlab{}.
\newblock \showarticletitle{AI chatbots for mental health: A scoping review of
  effectiveness, feasibility, and applications}.
\newblock \bibinfo{journal}{\emph{Appl. Sci}}  \bibinfo{volume}{14}
  (\bibinfo{year}{2024}), \bibinfo{pages}{5889}.
\newblock


\bibitem[Cegin et~al\mbox{.}(2024)]%
        {cegin2024llms}
\bibfield{author}{\bibinfo{person}{Jan Cegin}, \bibinfo{person}{Jakub Simko},
  {and} \bibinfo{person}{Peter Brusilovsky}.} \bibinfo{year}{2024}\natexlab{}.
\newblock \showarticletitle{LLMs vs Established Text Augmentation Techniques
  for Classification: When do the Benefits Outweight the Costs?}
\newblock \bibinfo{journal}{\emph{arXiv preprint arXiv:2408.16502}}
  (\bibinfo{year}{2024}).
\newblock


\bibitem[Dayanand et~al\mbox{.}({[n.\,d.]})]%
        {dayanandharnessing}
\bibfield{author}{\bibinfo{person}{Ananya~Kumari Dayanand},
  \bibinfo{person}{Rohit Tanwar}, {and} \bibinfo{person}{Shahina Anwarul}.}
  \bibinfo{year}{[n.\,d.]}\natexlab{}.
\newblock \showarticletitle{Harnessing Emotion Detection in Healthcare:
  Techniques, Challenges, and Future Directions}.
\newblock \bibinfo{journal}{\emph{Sentiment Analysis Unveiled}}
  (\bibinfo{year}{[n.\,d.]}), \bibinfo{pages}{131--150}.
\newblock


\bibitem[Demszky et~al\mbox{.}(2020)]%
        {demszky2020goemotions}
\bibfield{author}{\bibinfo{person}{Dorottya Demszky}, \bibinfo{person}{Dana
  Movshovitz-Attias}, \bibinfo{person}{Jeongwoo Ko}, \bibinfo{person}{Alan
  Cowen}, \bibinfo{person}{Gaurav Nemade}, {and} \bibinfo{person}{Sujith
  Ravi}.} \bibinfo{year}{2020}\natexlab{}.
\newblock \showarticletitle{GoEmotions: A dataset of fine-grained emotions}.
\newblock \bibinfo{journal}{\emph{arXiv preprint arXiv:2005.00547}}
  (\bibinfo{year}{2020}).
\newblock


\bibitem[Duan et~al\mbox{.}(2024)]%
        {duan2024large}
\bibfield{author}{\bibinfo{person}{Yucong Duan}, \bibinfo{person}{Fuliang
  Tang}, \bibinfo{person}{Kunguang Wu}, \bibinfo{person}{Zhendong Guo},
  \bibinfo{person}{Shuaishuai Huang}, \bibinfo{person}{Yingtian Mei},
  \bibinfo{person}{Yuxing Wang}, \bibinfo{person}{Zeyu Yang}, {and}
  \bibinfo{person}{Shiming Gong}.} \bibinfo{year}{2024}\natexlab{}.
\newblock \showarticletitle{The large language model (llm) bias evaluation (age
  bias)}.
\newblock \bibinfo{journal}{\emph{DIKWP Research Group International Standard
  Evaluation. DOI}}  \bibinfo{volume}{10} (\bibinfo{year}{2024}).
\newblock


\bibitem[Elyasi et~al\mbox{.}(2023)]%
        {elyasi2023exploring}
\bibfield{author}{\bibinfo{person}{Soroush Elyasi},
  \bibinfo{person}{Arya~Varasteh Nezhad}, {and} \bibinfo{person}{Fattaneh
  Taghiyareh}.} \bibinfo{year}{2023}\natexlab{}.
\newblock \showarticletitle{Exploring the Relationship Between Gameplay Log
  Data and Depression \& Anxiety}. In \bibinfo{booktitle}{\emph{2023 14th
  International Conference on Information and Knowledge Technology (IKT)}}.
  IEEE, \bibinfo{pages}{50--56}.
\newblock


\bibitem[Elyasi and Taghiyareh(2023)]%
        {elyasi2023mbti}
\bibfield{author}{\bibinfo{person}{Soroush Elyasi} {and}
  \bibinfo{person}{Fattaneh Taghiyareh}.} \bibinfo{year}{2023}\natexlab{}.
\newblock \showarticletitle{MBTI-Based Personality Assessment through
  Introducing a Puzzle Game}. In \bibinfo{booktitle}{\emph{2023 9th
  International Conference on Web Research (ICWR)}}. IEEE,
  \bibinfo{pages}{102--107}.
\newblock


\bibitem[Elyasi et~al\mbox{.}(2025)]%
        {elyasi2025play}
\bibfield{author}{\bibinfo{person}{Soroush Elyasi}, \bibinfo{person}{Arya
  VarastehNezhad}, {and} \bibinfo{person}{Fattaneh Taghiyareh}.}
  \bibinfo{year}{2025}\natexlab{}.
\newblock \showarticletitle{From Play to Prediction: Assessing Depression and
  Anxiety in Players Behavior with Machine Learning Models}.
\newblock \bibinfo{journal}{\emph{International Journal of Serious Games}}
  \bibinfo{volume}{12}, \bibinfo{number}{1} (\bibinfo{year}{2025}),
  \bibinfo{pages}{83--102}.
\newblock


\bibitem[Ettman and Galea(2023)]%
        {ettman2023potential}
\bibfield{author}{\bibinfo{person}{Catherine~K Ettman} {and}
  \bibinfo{person}{Sandro Galea}.} \bibinfo{year}{2023}\natexlab{}.
\newblock \showarticletitle{The potential influence of AI on population mental
  health}.
\newblock \bibinfo{journal}{\emph{JMIR Mental Health}}  \bibinfo{volume}{10}
  (\bibinfo{year}{2023}), \bibinfo{pages}{e49936}.
\newblock


\bibitem[Farhat(2024)]%
        {farhat2024chatgpt}
\bibfield{author}{\bibinfo{person}{Faiza Farhat}.}
  \bibinfo{year}{2024}\natexlab{}.
\newblock \showarticletitle{ChatGPT as a complementary mental health resource:
  a boon or a bane}.
\newblock \bibinfo{journal}{\emph{Annals of Biomedical Engineering}}
  \bibinfo{volume}{52}, \bibinfo{number}{5} (\bibinfo{year}{2024}),
  \bibinfo{pages}{1111--1114}.
\newblock


\bibitem[Fawaz and Samaha(2021)]%
        {fawaz2021}
\bibfield{author}{\bibinfo{person}{Mirna Fawaz} {and} \bibinfo{person}{Ali
  Samaha}.} \bibinfo{year}{2021}\natexlab{}.
\newblock \showarticletitle{E-learning: Depression, anxiety, and stress
  symptomatology among Lebanese university students during COVID-19
  quarantine}.
\newblock \bibinfo{journal}{\emph{Nursing Forum}} \bibinfo{volume}{56},
  \bibinfo{number}{1} (\bibinfo{year}{2021}), \bibinfo{pages}{52--57}.
\newblock


\bibitem[Friedrich(2017)]%
        {friedrich2017}
\bibfield{author}{\bibinfo{person}{Mary~Jane Friedrich}.}
  \bibinfo{year}{2017}\natexlab{}.
\newblock \showarticletitle{Depression is the leading cause of disability
  around the world}.
\newblock \bibinfo{journal}{\emph{Jama}} \bibinfo{volume}{317},
  \bibinfo{number}{15} (\bibinfo{year}{2017}), \bibinfo{pages}{1517--1517}.
\newblock
\href{https://doi.org/10.1001/jama.2017.3826}{doi:\nolinkurl{10.1001/jama.2017.3826}}


\bibitem[Gallegos et~al\mbox{.}(2024)]%
        {gallegos2024bias}
\bibfield{author}{\bibinfo{person}{Isabel~O Gallegos}, \bibinfo{person}{Ryan~A
  Rossi}, \bibinfo{person}{Joe Barrow}, \bibinfo{person}{Md~Mehrab Tanjim},
  \bibinfo{person}{Sungchul Kim}, \bibinfo{person}{Franck Dernoncourt},
  \bibinfo{person}{Tong Yu}, \bibinfo{person}{Ruiyi Zhang}, {and}
  \bibinfo{person}{Nesreen~K Ahmed}.} \bibinfo{year}{2024}\natexlab{}.
\newblock \showarticletitle{Bias and fairness in large language models: A
  survey}.
\newblock \bibinfo{journal}{\emph{Computational Linguistics}}
  \bibinfo{volume}{50}, \bibinfo{number}{3} (\bibinfo{year}{2024}),
  \bibinfo{pages}{1097--1179}.
\newblock


\bibitem[Health(2020)]%
        {health2020mental}
\bibfield{author}{\bibinfo{person}{The Lancet~Global Health}.}
  \bibinfo{year}{2020}\natexlab{}.
\newblock \showarticletitle{Mental health matters}.
\newblock \bibinfo{journal}{\emph{The Lancet. Global Health}}
  \bibinfo{volume}{8}, \bibinfo{number}{11} (\bibinfo{year}{2020}),
  \bibinfo{pages}{e1352}.
\newblock


\bibitem[Hua et~al\mbox{.}(2024)]%
        {hua2024large}
\bibfield{author}{\bibinfo{person}{Yining Hua}, \bibinfo{person}{Fenglin Liu},
  \bibinfo{person}{Kailai Yang}, \bibinfo{person}{Zehan Li},
  \bibinfo{person}{Hongbin Na}, \bibinfo{person}{Yi-han Sheu},
  \bibinfo{person}{Peilin Zhou}, \bibinfo{person}{Lauren~V Moran},
  \bibinfo{person}{Sophia Ananiadou}, {and} \bibinfo{person}{Andrew Beam}.}
  \bibinfo{year}{2024}\natexlab{}.
\newblock \showarticletitle{Large language models in mental health care: a
  scoping review}.
\newblock \bibinfo{journal}{\emph{arXiv preprint arXiv:2401.02984}}
  (\bibinfo{year}{2024}).
\newblock


\bibitem[Huang et~al\mbox{.}(2019)]%
        {huang2019clinicalbert}
\bibfield{author}{\bibinfo{person}{Kexin Huang}, \bibinfo{person}{Jaan
  Altosaar}, {and} \bibinfo{person}{Rajesh Ranganath}.}
  \bibinfo{year}{2019}\natexlab{}.
\newblock \showarticletitle{Clinicalbert: Modeling clinical notes and
  predicting hospital readmission}.
\newblock \bibinfo{journal}{\emph{arXiv preprint arXiv:1904.05342}}
  (\bibinfo{year}{2019}).
\newblock


\bibitem[Ilias et~al\mbox{.}(2023)]%
        {ilias2023calibration}
\bibfield{author}{\bibinfo{person}{Loukas Ilias}, \bibinfo{person}{Spiros
  Mouzakitis}, {and} \bibinfo{person}{Dimitris Askounis}.}
  \bibinfo{year}{2023}\natexlab{}.
\newblock \showarticletitle{Calibration of transformer-based models for
  identifying stress and depression in social media}.
\newblock \bibinfo{journal}{\emph{IEEE Transactions on Computational Social
  Systems}} \bibinfo{volume}{11}, \bibinfo{number}{2} (\bibinfo{year}{2023}),
  \bibinfo{pages}{1979--1990}.
\newblock


\bibitem[Imran et~al\mbox{.}(2023)]%
        {imran2023chat}
\bibfield{author}{\bibinfo{person}{Nazish Imran}, \bibinfo{person}{Aateqa
  Hashmi}, {and} \bibinfo{person}{Ahad Imran}.}
  \bibinfo{year}{2023}\natexlab{}.
\newblock \showarticletitle{Chat-GPT: opportunities and challenges in child
  mental healthcare}.
\newblock \bibinfo{journal}{\emph{Pakistan Journal of Medical Sciences}}
  \bibinfo{volume}{39}, \bibinfo{number}{4} (\bibinfo{year}{2023}),
  \bibinfo{pages}{1191}.
\newblock


\bibitem[Jin et~al\mbox{.}(2025)]%
        {jin2025applications}
\bibfield{author}{\bibinfo{person}{Yu Jin}, \bibinfo{person}{Jiayi Liu},
  \bibinfo{person}{Pan Li}, \bibinfo{person}{Baosen Wang},
  \bibinfo{person}{Yangxinyu Yan}, \bibinfo{person}{Huilin Zhang},
  \bibinfo{person}{Chenhao Ni}, \bibinfo{person}{Jing Wang},
  \bibinfo{person}{Yi Li}, {and} \bibinfo{person}{Yajun Bu}.}
  \bibinfo{year}{2025}\natexlab{}.
\newblock \showarticletitle{The Applications of Large Language Models in Mental
  Health: Scoping Review}.
\newblock \bibinfo{journal}{\emph{Journal of Medical Internet Research}}
  \bibinfo{volume}{27} (\bibinfo{year}{2025}), \bibinfo{pages}{e69284}.
\newblock


\bibitem[Kamarunzaman et~al\mbox{.}(2020)]%
        {kamarunzaman2020mental}
\bibfield{author}{\bibinfo{person}{Nur~Zafifa Kamarunzaman},
  \bibinfo{person}{Alice Shanthi}, \bibinfo{person}{Z Md~Nen},
  \bibinfo{person}{Norfarhana Zulkifli}, {and} \bibinfo{person}{Nur~Izzati
  Shamsuri}.} \bibinfo{year}{2020}\natexlab{}.
\newblock \showarticletitle{Mental health issues among university students and
  educators’ readiness to detect and help}.
\newblock \bibinfo{journal}{\emph{International Journal of Academic Research in
  Business and Social Sciences}} \bibinfo{volume}{10}, \bibinfo{number}{9}
  (\bibinfo{year}{2020}), \bibinfo{pages}{711--725}.
\newblock


\bibitem[Kleine et~al\mbox{.}(2023)]%
        {kleine2023attitudes}
\bibfield{author}{\bibinfo{person}{Anne-Kathrin Kleine}, \bibinfo{person}{Eesha
  Kokje}, \bibinfo{person}{Eva Lermer}, {and} \bibinfo{person}{Susanne Gaube}.}
  \bibinfo{year}{2023}\natexlab{}.
\newblock \showarticletitle{Attitudes toward the adoption of 2 artificial
  intelligence--enabled mental health tools among prospective psychotherapists:
  Cross-sectional study}.
\newblock \bibinfo{journal}{\emph{JMIR human factors}}  \bibinfo{volume}{10}
  (\bibinfo{year}{2023}), \bibinfo{pages}{e46859}.
\newblock


\bibitem[Kour and Gupta(2022)]%
        {kour2022hybrid}
\bibfield{author}{\bibinfo{person}{Harnain Kour} {and} \bibinfo{person}{Manoj~K
  Gupta}.} \bibinfo{year}{2022}\natexlab{}.
\newblock \showarticletitle{An hybrid deep learning approach for depression
  prediction from user tweets using feature-rich CNN and bi-directional LSTM}.
\newblock \bibinfo{journal}{\emph{Multimedia Tools and Applications}}
  \bibinfo{volume}{81}, \bibinfo{number}{17} (\bibinfo{year}{2022}),
  \bibinfo{pages}{23649--23685}.
\newblock


\bibitem[Lahoti et~al\mbox{.}(2023)]%
        {lahoti2023improving}
\bibfield{author}{\bibinfo{person}{Preethi Lahoti}, \bibinfo{person}{Nicholas
  Blumm}, \bibinfo{person}{Xiao Ma}, \bibinfo{person}{Raghavendra
  Kotikalapudi}, \bibinfo{person}{Sahitya Potluri}, \bibinfo{person}{Qijun
  Tan}, \bibinfo{person}{Hansa Srinivasan}, \bibinfo{person}{Ben Packer},
  \bibinfo{person}{Ahmad Beirami}, {and} \bibinfo{person}{Alex Beutel}.}
  \bibinfo{year}{2023}\natexlab{}.
\newblock \showarticletitle{Improving diversity of demographic representation
  in large language models via collective-critiques and self-voting}.
\newblock \bibinfo{journal}{\emph{arXiv preprint arXiv:2310.16523}}
  (\bibinfo{year}{2023}).
\newblock


\bibitem[Lee et~al\mbox{.}(2020)]%
        {lee2020biobert}
\bibfield{author}{\bibinfo{person}{Jinhyuk Lee}, \bibinfo{person}{Wonjin Yoon},
  \bibinfo{person}{Sungdong Kim}, \bibinfo{person}{Donghyeon Kim},
  \bibinfo{person}{Sunkyu Kim}, \bibinfo{person}{Chan~Ho So}, {and}
  \bibinfo{person}{Jaewoo Kang}.} \bibinfo{year}{2020}\natexlab{}.
\newblock \showarticletitle{BioBERT: a pre-trained biomedical language
  representation model for biomedical text mining}.
\newblock \bibinfo{journal}{\emph{Bioinformatics}} \bibinfo{volume}{36},
  \bibinfo{number}{4} (\bibinfo{year}{2020}), \bibinfo{pages}{1234--1240}.
\newblock


\bibitem[Lissak et~al\mbox{.}(2024)]%
        {lissak2024colorful}
\bibfield{author}{\bibinfo{person}{Shir Lissak}, \bibinfo{person}{Nitay
  Calderon}, \bibinfo{person}{Geva Shenkman}, \bibinfo{person}{Yaakov Ophir},
  \bibinfo{person}{Eyal Fruchter}, \bibinfo{person}{Anat~Brunstein Klomek},
  {and} \bibinfo{person}{Roi Reichart}.} \bibinfo{year}{2024}\natexlab{}.
\newblock \showarticletitle{The colorful future of llms: Evaluating and
  improving llms as emotional supporters for queer youth}.
\newblock \bibinfo{journal}{\emph{arXiv preprint arXiv:2402.11886}}
  (\bibinfo{year}{2024}).
\newblock


\bibitem[Ma et~al\mbox{.}(2024)]%
        {ma2024understanding}
\bibfield{author}{\bibinfo{person}{Zilin Ma}, \bibinfo{person}{Yiyang Mei},
  {and} \bibinfo{person}{Zhaoyuan Su}.} \bibinfo{year}{2024}\natexlab{}.
\newblock \showarticletitle{Understanding the benefits and challenges of using
  large language model-based conversational agents for mental well-being
  support}. In \bibinfo{booktitle}{\emph{AMIA Annual Symposium Proceedings}},
  Vol.~\bibinfo{volume}{2023}. \bibinfo{pages}{1105}.
\newblock


\bibitem[Mekonen et~al\mbox{.}(2021)]%
        {mekonen2021estimating}
\bibfield{author}{\bibinfo{person}{Tesfa Mekonen}, \bibinfo{person}{Gary~CK
  Chan}, \bibinfo{person}{Jason~P Connor}, \bibinfo{person}{Leanne Hides},
  {and} \bibinfo{person}{Janni Leung}.} \bibinfo{year}{2021}\natexlab{}.
\newblock \showarticletitle{Estimating the global treatment rates for
  depression: a systematic review and meta-analysis}.
\newblock \bibinfo{journal}{\emph{Journal of Affective Disorders}}
  \bibinfo{volume}{295} (\bibinfo{year}{2021}), \bibinfo{pages}{1234--1242}.
\newblock


\bibitem[Mendel et~al\mbox{.}(2025)]%
        {mendel2025laypeople}
\bibfield{author}{\bibinfo{person}{Tamir Mendel}, \bibinfo{person}{Nina Singh},
  \bibinfo{person}{Devin~M Mann}, \bibinfo{person}{Batia Wiesenfeld}, {and}
  \bibinfo{person}{Oded Nov}.} \bibinfo{year}{2025}\natexlab{}.
\newblock \showarticletitle{Laypeople’s Use of and Attitudes Toward Large
  Language Models and Search Engines for Health Queries: Survey Study}.
\newblock \bibinfo{journal}{\emph{Journal of Medical Internet Research}}
  \bibinfo{volume}{27} (\bibinfo{year}{2025}), \bibinfo{pages}{e64290}.
\newblock


\bibitem[Minerva and Giubilini(2023)]%
        {minerva2023ai}
\bibfield{author}{\bibinfo{person}{Francesca Minerva} {and}
  \bibinfo{person}{Alberto Giubilini}.} \bibinfo{year}{2023}\natexlab{}.
\newblock \showarticletitle{Is AI the future of mental healthcare?}
\newblock \bibinfo{journal}{\emph{Topoi}} \bibinfo{volume}{42},
  \bibinfo{number}{3} (\bibinfo{year}{2023}), \bibinfo{pages}{809--817}.
\newblock


\bibitem[Nguyen et~al\mbox{.}(2024)]%
        {nguyen2024sentiment}
\bibfield{author}{\bibinfo{person}{Khai-Nguyen Nguyen}, \bibinfo{person}{Khai
  Le-Duc}, \bibinfo{person}{Bach~Phan Tat}, \bibinfo{person}{Duy Le},
  \bibinfo{person}{Long Vo-Dang}, {and} \bibinfo{person}{Truong-Son Hy}.}
  \bibinfo{year}{2024}\natexlab{}.
\newblock \showarticletitle{Sentiment Reasoning for Healthcare}.
\newblock \bibinfo{journal}{\emph{arXiv preprint arXiv:2407.21054}}
  (\bibinfo{year}{2024}).
\newblock


\bibitem[Niriella et~al\mbox{.}(2025)]%
        {niriella2025reliability}
\bibfield{author}{\bibinfo{person}{Madunil~A Niriella}, \bibinfo{person}{Pathum
  Premaratna}, \bibinfo{person}{Mananjala Senanayake},
  \bibinfo{person}{Senerath Kodisinghe}, \bibinfo{person}{Uditha Dassanayake},
  \bibinfo{person}{Anuradha Dassanayake}, \bibinfo{person}{Dileepa~S
  Ediriweera}, {and} \bibinfo{person}{H~Janaka de Silva}.}
  \bibinfo{year}{2025}\natexlab{}.
\newblock \showarticletitle{The reliability of freely accessible, baseline,
  general-purpose large language model generated patient information for
  frequently asked questions on liver disease: a preliminary cross-sectional
  study}.
\newblock \bibinfo{journal}{\emph{Expert Review of Gastroenterology \&
  Hepatology}} \bibinfo{volume}{19}, \bibinfo{number}{4}
  (\bibinfo{year}{2025}), \bibinfo{pages}{437--442}.
\newblock


\bibitem[Obradovich et~al\mbox{.}(2024)]%
        {obradovich2024opportunities}
\bibfield{author}{\bibinfo{person}{Nick Obradovich}, \bibinfo{person}{Sahib~S
  Khalsa}, \bibinfo{person}{Waqas~U Khan}, \bibinfo{person}{Jina Suh},
  \bibinfo{person}{Roy~H Perlis}, \bibinfo{person}{Olusola Ajilore}, {and}
  \bibinfo{person}{Martin~P Paulus}.} \bibinfo{year}{2024}\natexlab{}.
\newblock \showarticletitle{Opportunities and risks of large language models in
  psychiatry}.
\newblock \bibinfo{journal}{\emph{NPP—Digital Psychiatry and Neuroscience}}
  \bibinfo{volume}{2}, \bibinfo{number}{1} (\bibinfo{year}{2024}),
  \bibinfo{pages}{8}.
\newblock


\bibitem[Ohse et~al\mbox{.}(2024)]%
        {ohse2024gpt}
\bibfield{author}{\bibinfo{person}{Julia Ohse}, \bibinfo{person}{Bakir
  Had{\v{z}}i{\'c}}, \bibinfo{person}{Parvez Mohammed},
  \bibinfo{person}{Nicolina Peperkorn}, \bibinfo{person}{Janosch Fox},
  \bibinfo{person}{Joshua Krutzki}, \bibinfo{person}{Alexander Lyko},
  \bibinfo{person}{Fan Mingyu}, \bibinfo{person}{Xiaohu Zheng}, {and}
  \bibinfo{person}{Matthias R{\"a}tsch}.} \bibinfo{year}{2024}\natexlab{}.
\newblock \showarticletitle{GPT-4 shows potential for identifying social
  anxiety from clinical interview data}.
\newblock \bibinfo{journal}{\emph{Scientific Reports}} \bibinfo{volume}{14},
  \bibinfo{number}{1} (\bibinfo{year}{2024}), \bibinfo{pages}{1--12}.
\newblock


\bibitem[Olawade et~al\mbox{.}(2024)]%
        {olawade2024enhancing}
\bibfield{author}{\bibinfo{person}{David~B Olawade}, \bibinfo{person}{Ojima~Z
  Wada}, \bibinfo{person}{Aderonke Odetayo},
  \bibinfo{person}{Aanuoluwapo~Clement David-Olawade},
  \bibinfo{person}{Fiyinfoluwa Asaolu}, {and} \bibinfo{person}{Judith
  Eberhardt}.} \bibinfo{year}{2024}\natexlab{}.
\newblock \showarticletitle{Enhancing mental health with Artificial
  Intelligence: Current trends and future prospects}.
\newblock \bibinfo{journal}{\emph{Journal of medicine, surgery, and public
  health}} (\bibinfo{year}{2024}), \bibinfo{pages}{100099}.
\newblock


\bibitem[Omiye et~al\mbox{.}(2024)]%
        {omiye2024large}
\bibfield{author}{\bibinfo{person}{Jesutofunmi~A Omiye},
  \bibinfo{person}{Haiwen Gui}, \bibinfo{person}{Shawheen~J Rezaei},
  \bibinfo{person}{James Zou}, {and} \bibinfo{person}{Roxana Daneshjou}.}
  \bibinfo{year}{2024}\natexlab{}.
\newblock \showarticletitle{Large language models in medicine: the potentials
  and pitfalls: a narrative review}.
\newblock \bibinfo{journal}{\emph{Annals of internal medicine}}
  \bibinfo{volume}{177}, \bibinfo{number}{2} (\bibinfo{year}{2024}),
  \bibinfo{pages}{210--220}.
\newblock


\bibitem[{OpenAI}(2025)]%
        {OAI_sycophancy2025}
\bibfield{author}{\bibinfo{person}{{OpenAI}}.} \bibinfo{year}{2025}\natexlab{}.
\newblock \bibinfo{title}{Expanding on what we missed with sycophancy}.
\newblock
  \bibinfo{howpublished}{\url{https://www.who.int/news-room/fact-sheets/detail/depression}}.
\newblock
\newblock
\shownote{Accessed: 2025-07-07}.


\bibitem[Palmer et~al\mbox{.}(2024)]%
        {palmer2024combining}
\bibfield{author}{\bibinfo{person}{Clare~E Palmer}, \bibinfo{person}{Emily
  Marshall}, \bibinfo{person}{Edward Millgate}, \bibinfo{person}{Graham
  Warren}, \bibinfo{person}{Michael~P Ewbank}, \bibinfo{person}{Elisa Cooper},
  \bibinfo{person}{Samantha Lawes}, \bibinfo{person}{Malika Bouazzaoui},
  \bibinfo{person}{Alastair Smith}, {and} \bibinfo{person}{Chris
  Hutchins-Joss}.} \bibinfo{year}{2024}\natexlab{}.
\newblock \showarticletitle{Combining AI and human support in mental health: A
  digital intervention with comparable effectiveness to human-delivered care}.
\newblock \bibinfo{journal}{\emph{medRxiv}} (\bibinfo{year}{2024}),
  \bibinfo{pages}{2024--07}.
\newblock


\bibitem[Park et~al\mbox{.}(2024)]%
        {park2024effectiveness}
\bibfield{author}{\bibinfo{person}{Yoonseo Park}, \bibinfo{person}{Sewon Park},
  {and} \bibinfo{person}{Munjae Lee}.} \bibinfo{year}{2024}\natexlab{}.
\newblock \showarticletitle{Effectiveness of artificial intelligence in
  detecting and managing depressive disorders: Systematic literature review}.
\newblock \bibinfo{journal}{\emph{Journal of Affective Disorders}}
  (\bibinfo{year}{2024}).
\newblock


\bibitem[Rojas et~al\mbox{.}(2025)]%
        {rojas2025health}
\bibfield{author}{\bibinfo{person}{Natalia~K Rojas}, \bibinfo{person}{Sam
  Martin}, \bibinfo{person}{Mario Cortina-Borja}, \bibinfo{person}{Roz
  Shafran}, \bibinfo{person}{Lana Fox-Smith}, \bibinfo{person}{Terence
  Stephenson}, \bibinfo{person}{Brian~CF Ching}, \bibinfo{person}{Ana{\"\i}s
  d'Oelsnitz}, \bibinfo{person}{Tom Norris}, {and} \bibinfo{person}{Yue Xu}.}
  \bibinfo{year}{2025}\natexlab{}.
\newblock \showarticletitle{Health and Experiences During the COVID-19 Pandemic
  Among Children and Young People: Analysis of Free-Text Responses From the
  Children and Young People With Long COVID Study}.
\newblock \bibinfo{journal}{\emph{Journal of Medical Internet Research}}
  \bibinfo{volume}{27} (\bibinfo{year}{2025}), \bibinfo{pages}{e63634}.
\newblock


\bibitem[Salinas et~al\mbox{.}(2023)]%
        {salinas2023unequal}
\bibfield{author}{\bibinfo{person}{Abel Salinas}, \bibinfo{person}{Parth Shah},
  \bibinfo{person}{Yuzhong Huang}, \bibinfo{person}{Robert McCormack}, {and}
  \bibinfo{person}{Fred Morstatter}.} \bibinfo{year}{2023}\natexlab{}.
\newblock \showarticletitle{The unequal opportunities of large language models:
  Examining demographic biases in job recommendations by chatgpt and llama}. In
  \bibinfo{booktitle}{\emph{Proceedings of the 3rd ACM Conference on Equity and
  Access in Algorithms, Mechanisms, and Optimization}}. \bibinfo{pages}{1--15}.
\newblock


\bibitem[Shen et~al\mbox{.}(2024)]%
        {shen2024chatgpt}
\bibfield{author}{\bibinfo{person}{Sarek~A Shen}, \bibinfo{person}{Carlos~A
  Perez-Heydrich}, \bibinfo{person}{Deborah~X Xie}, {and}
  \bibinfo{person}{Jason~C Nellis}.} \bibinfo{year}{2024}\natexlab{}.
\newblock \showarticletitle{ChatGPT vs. web search for patient questions: what
  does ChatGPT do better?}
\newblock \bibinfo{journal}{\emph{European Archives of Oto-Rhino-Laryngology}}
  \bibinfo{volume}{281}, \bibinfo{number}{6} (\bibinfo{year}{2024}),
  \bibinfo{pages}{3219--3225}.
\newblock


\bibitem[Tavasoli et~al\mbox{.}(2025)]%
        {tavasoli2025analyzing}
\bibfield{author}{\bibinfo{person}{Reza Tavasoli}, \bibinfo{person}{Arya
  VarastehNezhad}, \bibinfo{person}{Mostafa Masumi}, {and}
  \bibinfo{person}{Fattaneh Taghiyareh}.} \bibinfo{year}{2025}\natexlab{}.
\newblock \showarticletitle{Analyzing the Mathematical Proficiency of Large
  Language Models in Computer Science Graduate Admission Tests}. In
  \bibinfo{booktitle}{\emph{2025 29th International Computer Conference,
  Computer Society of Iran (CSICC)}}. IEEE, \bibinfo{pages}{1--5}.
\newblock


\bibitem[Tejaswini et~al\mbox{.}(2024)]%
        {tejaswini2024depression}
\bibfield{author}{\bibinfo{person}{Vankayala Tejaswini}, \bibinfo{person}{Korra
  Sathya~Babu}, {and} \bibinfo{person}{Bibhudatta Sahoo}.}
  \bibinfo{year}{2024}\natexlab{}.
\newblock \showarticletitle{Depression detection from social media text
  analysis using natural language processing techniques and hybrid deep
  learning model}.
\newblock \bibinfo{journal}{\emph{ACM Transactions on Asian and Low-Resource
  Language Information Processing}} \bibinfo{volume}{23}, \bibinfo{number}{1}
  (\bibinfo{year}{2024}), \bibinfo{pages}{1--20}.
\newblock


\bibitem[Thirunavukarasu et~al\mbox{.}(2023)]%
        {thirunavukarasu2023large}
\bibfield{author}{\bibinfo{person}{Arun~James Thirunavukarasu},
  \bibinfo{person}{Darren Shu~Jeng Ting}, \bibinfo{person}{Kabilan Elangovan},
  \bibinfo{person}{Laura Gutierrez}, \bibinfo{person}{Ting~Fang Tan}, {and}
  \bibinfo{person}{Daniel Shu~Wei Ting}.} \bibinfo{year}{2023}\natexlab{}.
\newblock \showarticletitle{Large language models in medicine}.
\newblock \bibinfo{journal}{\emph{Nature medicine}} \bibinfo{volume}{29},
  \bibinfo{number}{8} (\bibinfo{year}{2023}), \bibinfo{pages}{1930--1940}.
\newblock


\bibitem[VarastehNezhad et~al\mbox{.}(2024b)]%
        {varastehnezhad2024evaluating}
\bibfield{author}{\bibinfo{person}{Arya VarastehNezhad}, \bibinfo{person}{Reza
  Tavasoli}, \bibinfo{person}{Mostafa Masumi}, \bibinfo{person}{Seyed~Soroush
  Majd}, {and} \bibinfo{person}{Mehrnoush Shamsfard}.}
  \bibinfo{year}{2024}\natexlab{b}.
\newblock \showarticletitle{Evaluating LLMs in Persian News Summarization}. In
  \bibinfo{booktitle}{\emph{2024 15th International Conference on Information
  and Knowledge Technology (IKT)}}. IEEE, \bibinfo{pages}{195--201}.
\newblock


\bibitem[VarastehNezhad et~al\mbox{.}(2024a)]%
        {varastehnezhad2024llm}
\bibfield{author}{\bibinfo{person}{Arya VarastehNezhad}, \bibinfo{person}{Reza
  Tavasoli}, \bibinfo{person}{Mostafa Masumi}, {and} \bibinfo{person}{Fattaneh
  Taghiyareh}.} \bibinfo{year}{2024}\natexlab{a}.
\newblock \showarticletitle{LLM Performance Assessment in Computer Science
  Graduate Entrance Exams}. In \bibinfo{booktitle}{\emph{2024 11th
  International Symposium on Telecommunications (IST)}}. IEEE,
  \bibinfo{pages}{232--237}.
\newblock


\bibitem[Vzorinab et~al\mbox{.}(2024)]%
        {vzorinab2024emotional}
\bibfield{author}{\bibinfo{person}{Gleb~D Vzorinab}, \bibinfo{person}{Alexey~M
  Bukinichac}, \bibinfo{person}{Anna~V Sedykha}, \bibinfo{person}{Irina~I
  Vetrovab}, {and} \bibinfo{person}{Elena~A Sergienkob}.}
  \bibinfo{year}{2024}\natexlab{}.
\newblock \showarticletitle{The Emotional Intelligence of the GPT-4 Large
  Language Model}.
\newblock \bibinfo{journal}{\emph{Psychology in Russia: State of the art}}
  \bibinfo{volume}{17}, \bibinfo{number}{2} (\bibinfo{year}{2024}),
  \bibinfo{pages}{85--99}.
\newblock


\bibitem[{World Health Organization}(2023)]%
        {WHO_Depression2023}
\bibfield{author}{\bibinfo{person}{{World Health Organization}}.}
  \bibinfo{year}{2023}\natexlab{}.
\newblock \bibinfo{title}{Depressive disorder (depression)}.
\newblock
  \bibinfo{howpublished}{\url{https://www.who.int/news-room/fact-sheets/detail/depression}}.
\newblock


\bibitem[{World Health Organization}(2023)]%
        {WHO_Anxiety2023}
\bibfield{author}{\bibinfo{person}{{World Health Organization}}.}
  \bibinfo{year}{2023}\natexlab{}.
\newblock \bibinfo{title}{Anxiety disorders}.
\newblock
  \bibinfo{howpublished}{\url{https://www.who.int/news-room/fact-sheets/detail/anxiety-disorders}}.
\newblock
\newblock
\shownote{Accessed: 2025-02-02}.


\bibitem[{World Health Organization}(2024)]%
        {NIH_Depression2023}
\bibfield{author}{\bibinfo{person}{{World Health Organization}}.}
  \bibinfo{year}{2024}\natexlab{}.
\newblock \bibinfo{title}{Anxiety disorders}.
\newblock
  \bibinfo{howpublished}{\url{https://www.nimh.nih.gov/health/topics/depression}}.
\newblock
\newblock
\shownote{Accessed: 2025-02-03}.


\bibitem[Xu et~al\mbox{.}(2023)]%
        {xu2023leveraging}
\bibfield{author}{\bibinfo{person}{Xuhai Xu}, \bibinfo{person}{Bingsheng Yao},
  \bibinfo{person}{Yuanzhe Dong}, \bibinfo{person}{Hong Yu},
  \bibinfo{person}{James~A Hendler}, \bibinfo{person}{Anind~K Dey}, {and}
  \bibinfo{person}{Dakuo Wang}.} \bibinfo{year}{2023}\natexlab{}.
\newblock \showarticletitle{Leveraging large language models for mental health
  prediction via online text data}.
\newblock  (\bibinfo{year}{2023}).
\newblock


\bibitem[Zhang et~al\mbox{.}(2022)]%
        {zhang2022natural}
\bibfield{author}{\bibinfo{person}{Tianlin Zhang}, \bibinfo{person}{Annika~M
  Schoene}, \bibinfo{person}{Shaoxiong Ji}, {and} \bibinfo{person}{Sophia
  Ananiadou}.} \bibinfo{year}{2022}\natexlab{}.
\newblock \showarticletitle{Natural language processing applied to mental
  illness detection: a narrative review}.
\newblock \bibinfo{journal}{\emph{NPJ digital medicine}} \bibinfo{volume}{5},
  \bibinfo{number}{1} (\bibinfo{year}{2022}), \bibinfo{pages}{46}.
\newblock


\bibitem[Zhou et~al\mbox{.}(2025)]%
        {zhou2025evaluating}
\bibfield{author}{\bibinfo{person}{Mi Zhou}, \bibinfo{person}{Xiaomei Song},
  \bibinfo{person}{Qin Hu}, {and} \bibinfo{person}{Youbin Zhou}.}
  \bibinfo{year}{2025}\natexlab{}.
\newblock \showarticletitle{Evaluating ChatGPT-4o’s Web-Enhanced Responses in
  Patient Education: Ankle Stabilization Surgery as a Case Study}.
\newblock  (\bibinfo{year}{2025}).
\newblock


\bibitem[Zucchetti et~al\mbox{.}(2024)]%
        {zucchetti2024artificial}
\bibfield{author}{\bibinfo{person}{Andrea Zucchetti}, \bibinfo{person}{Gabriele
  Nibbio}, \bibinfo{person}{Luca Altieri}, \bibinfo{person}{Lorenzo Bertorni},
  \bibinfo{person}{Irene Calzavara-Pinton}, \bibinfo{person}{Elena Invernizzi},
  \bibinfo{person}{Nicola Necchini}, \bibinfo{person}{Caterina Cerati},
  \bibinfo{person}{Laura Poddighe}, {and} \bibinfo{person}{Viola Bulgari}.}
  \bibinfo{year}{2024}\natexlab{}.
\newblock \showarticletitle{Artificial intelligence applications in mental
  health: The state of the art}.
\newblock \bibinfo{journal}{\emph{Italian Journal of Psychiatry}}
  (\bibinfo{year}{2024}).
\newblock


\end{thebibliography}

\end{document}